\documentclass[11pt]{article}

\usepackage[final]{acl}

\usepackage{times}
\usepackage{latexsym}
\usepackage{amsmath, amssymb, amsfonts}
\usepackage{enumerate, enumitem}
\usepackage{parskip}
\usepackage{booktabs, multirow, adjustbox, makecell}
\usepackage{float}
\usepackage{array}
\usepackage[most]{tcolorbox}
\newcommand{\googlelogo}{%
    \raisebox{-0.15em}{\includegraphics[height=0.9em]{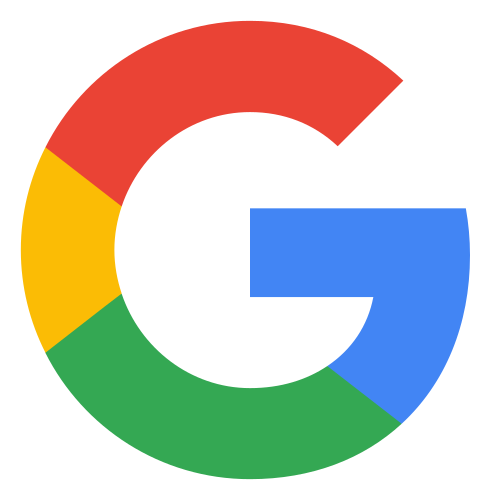}}%
}
\newcommand{\anthropiclogo}{%
    \raisebox{-0.15em}{\includegraphics[height=0.9em]{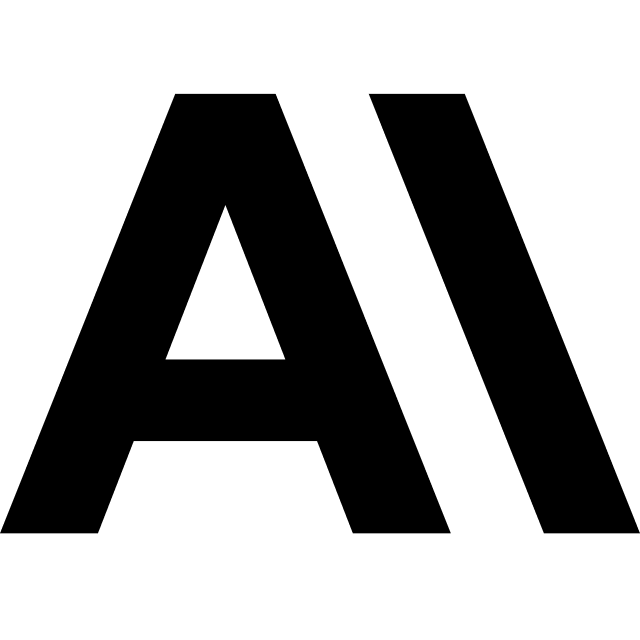}}%
}
\newcommand{\openailogo}{%
    \raisebox{-0.15em}{\includegraphics[height=0.9em]{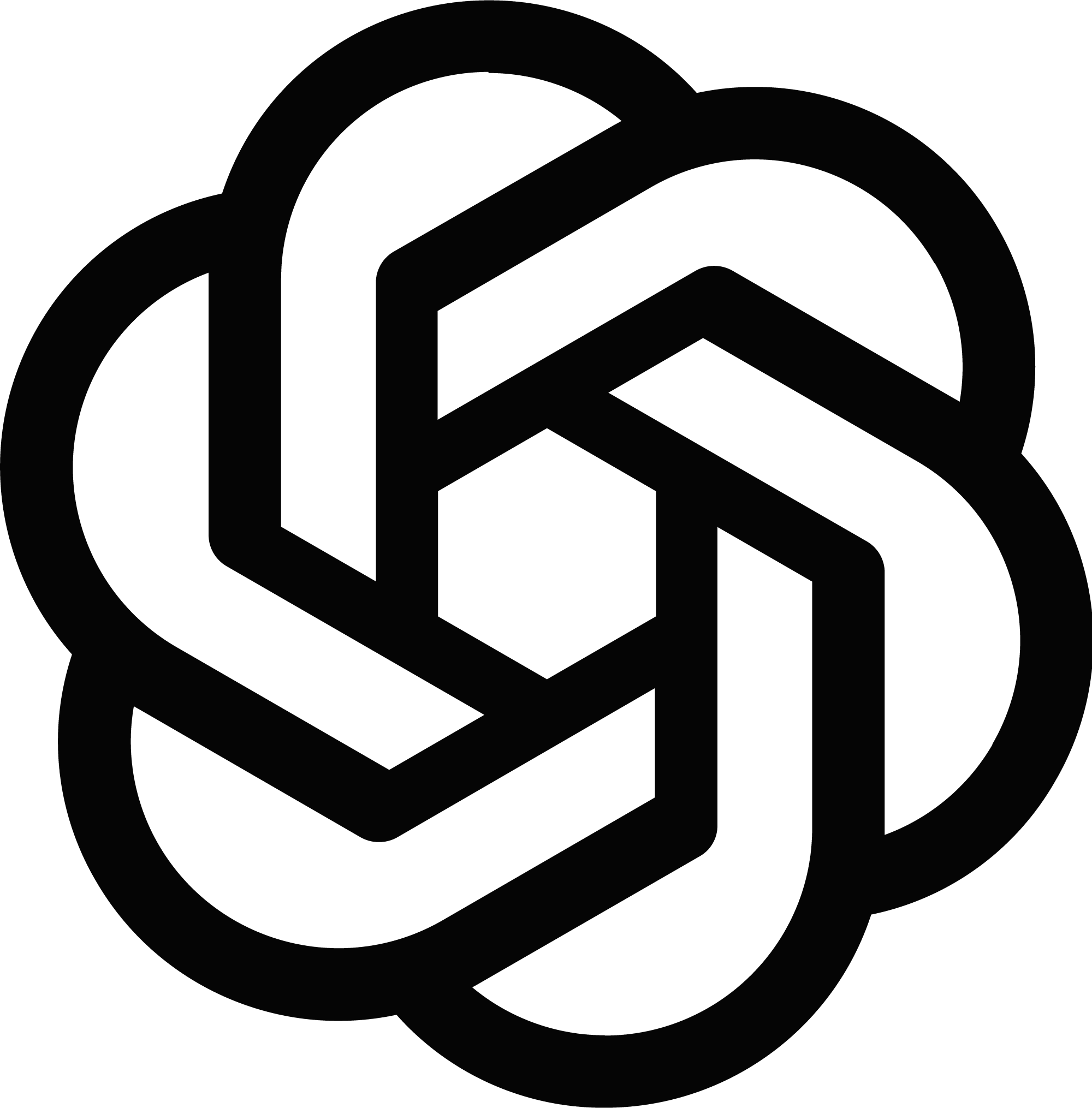}}%
}

\usepackage[T1]{fontenc}

\usepackage[utf8]{inputenc}

\usepackage{microtype}

\usepackage{inconsolata}

\usepackage{graphicx}

\title{Diagnose, Then Repair: A Two-Stage MQM-Guided \\ Post-Editing Framework for Domain-Specific Machine Translation}

\author{Ji Hun Wang \\
  Amazon \\
  \texttt{jihunw@amazon.com} \\\And
  Siyu Wu \\
  Amazon \\
  \texttt{siyumw@amazon.com} \\}

\begin{document}
\maketitle
\begin{abstract}
LLM-based machine translation evaluation can closely match human judgments, but in practice it remains largely diagnostic, with the signals rarely translating into direct quality improvements under real production constraints. We propose a two-stage, evaluator-guided automatic post-editing framework that turns MQM-style evaluation into targeted repairs: a retrieval-augmented LLM evaluator outputs structured, span-level MQM diagnoses under an explicit edit contract, and a separate LLM post-editor applies minimal edits restricted to those diagnoses. This separation improves controllability and reduces paraphrastic drift compared to one-stage ``judge-and-refine'' baselines. In a systematic study involving seven LLMs spanning three model providers and seven languages, our best configuration consistently improves both \textsc{Comet}-22 and \textsc{CometKiwi} scores over one-stage post-edit methods, while the evaluator's error spans and severities show strong agreement with human MQM annotations and human editor preferences.
\end{abstract}

\section{Introduction} \label{sec:introduction}

Machine Translation (MT) systems deployed in specialized domains face recurring failures driven by domain mismatch and domain-specific constraints such as terminology, rare or novel expressions, and style conventions \citep{koehn2017challengesneuralmachinetranslation}. In production settings, quality improvements must also crucially respect fixed compute and latency budgets and operational constraints, making frequent re-training or large-scale re-deployments challenging and difficult to justify economically. Domain distributions are also rarely static, for content refreshes and vocabulary churn can introduce new terms or deprecate old ones, which can quickly erode domain faithfulness even when general MT quality remains strong \citep{koehn2017challengesneuralmachinetranslation}.

Instruction-following LLMs have emerged as an effective solution in this context, for they can often adapt translation behavior via prompting with in-context learning (ICL) and demonstrate competitive translation quality \citep{jiao2023chatgptgoodtranslatoryes, alves-etal-2023-steering, wang-etal-2023-document-level}. However, domain MT still requires evaluation signals aligned with human expectations. Simple algorithmic scores, such as BLEU, are often insufficient and unreliable metrics, missing many adequacy- and terminology-related issues \citep{mathur-etal-2020-tangled}, while learned metrics such as \textsc{Comet}-22 correlate better with human judgments \citep{rei-etal-2020-comet} but inherently cannot localize what should be fixed.

This has motivated the use of descriptive rubrics, such as Multidimensional Quality Metrics (MQM) that provides a structured taxonomy with error category and severity, thereby enabling interpretable error analysis and actionable diagnosis \citep{burchardt-2013-multidimensional, lommel-etal-2014-using}. Recent work has shown that LLMs can be prompted to produce MQM-style error spans directly \citep{kocmi2023gembamqmdetectingtranslationquality} and that rubric-style prompting can further improve span-level LLM judgment for high-end models \citep{kim-2025-rubric}. Moreover, MQM-derived feedback can be used to guide LLM post-editing of MT outputs \citep{ki-carpuat-2024-guiding, madaan2023selfrefineiterativerefinementselffeedback}.

A gap in this thread of research is that self-refinement is difficult to control, and that the resulting gains could often be inconsistent depending on the models and sensitive to prompt design and task setup \citep{madaan2023selfrefineiterativerefinementselffeedback}. Furthermore, studies of intrinsic self-correction find that LLMs often fail to reliably detect and fix their own mistakes without external feedback, and repeated self-correction can degrade performance \citep{huang2024largelanguagemodelsselfcorrect}.

To that end, we propose a modified two-stage MT output repair framework that separates post-edit LLMs from evaluator LLMs, preserving the benefits of MQM-style annotations and error analysis while allowing for directly targeted, controlled fixes while circumventing LLM self-correction.

Our key contributions are as follows:
\begin{itemize}
    \item We propose a \textbf{two-stage MT post-edit framework} that evaluates first and then post-edits. We empirically validate its effectiveness compared to one-stage approach. 
    \item We investigate how different prompting strategies could affect post-edit performance, and show that our prompt that incorporates RAG and ``edit contract'' (Section~\ref{subsec:edit-contracts}) generates better post-edits.
    \item Through a systematic study involving seven models spanning three model providers and various model sizes, we assess how model diversity and model sizes could affect the post-edit performance, which could be used as general guidance when adopting the framework.
\end{itemize}

\section{Related Work} \label{sec:related-work}

\textbf{MQM-based evaluation and LLM annotators.} $\,$ MQM is a fine-grained framework for MT quality assessment that organizes errors into a flexible taxonomy including error labels and severity, making evaluation actionable beyond single-score metrics \citep{burchardt-2013-multidimensional, lommel-etal-2014-using}. It has also been used in large-scale expert human evaluation setups \citep{freitag-etal-2021-experts} and as a basis for meta-evaluation in WMT \citep{freitag-etal-2022-results}. ``LLM-as-a-judge or jury'' work has operationalized MQM with structured prompting, so that LLMs can produce rubric-faithful MQM judgments, including span-level error localizations and reasoning \citep{zheng2023judgingllmasajudgemtbenchchatbot, liu-etal-2023-g, chiang-lee-2023-closer, verga2024replacingjudgesjuriesevaluating}. Works such as GEMBA-MQM \citep{kocmi2023gembamqmdetectingtranslationquality} and \textsc{Rubric}-MQM \citep{kim-2025-rubric} build on this thread to make the approach more reliable and aligned with human judgments.


\textbf{Post-editing translations with LLMs.} $\,$ Automatic post-editing of MT outputs using strong instruction-following LLMs has garnered attention as an effective approach to directly improve the translations without re-training the MT model \citep{chatterjee-etal-2018-findings, simard-etal-2007-statistical}. It has been shown that a single-shot post-edit or iterative refinement of edits that alternate assessment and revision are both effective \citep{chen-etal-2024-iterative, wu2025translateagainsimpleexperiments, briakou2024translatingstepbystepdecomposingtranslation}. A closely related thread explicitly couples quality estimation or structured error feedback with post-editing, motivating pipelines where an evaluator's diagnostic signal guides subsequent edits rather than relying on generic instructions \citep{ki-carpuat-2024-guiding, lu2024mqmapehighqualityerrorannotation}.

\textbf{Retrieval-augmented techniques for domain MT.} $\,$ Retrieval-augmented generation (RAG) is a technique that incorporates relevant information from external sources to MT models (such as seq-2-seq models or LLMs) to boost their domain-specific knowledge \citep{lewis2021retrievalaugmentedgenerationknowledgeintensivenlp}. Existing work on domain MT has leveraged paired sentence exemplars retrieved from an external source as a method of data augmentation or decoding heuristics \citep{zhang-etal-2018-guiding, bulte-tezcan-2019-neural, khandelwal2021nearestneighbormachinetranslation} or a multilingual knowledge graphs using domain-specific examples \citep{conia-etal-2024-towards, wang2025retrievalaugmentedmachinetranslationunstructured}, to guide the model on terminological or stylistic elements specific to the domain.

\section{Method} \label{sec:method}

\subsection{Overview} \label{subsec:overview}

We study \textbf{segment-level domain-specific MT}. Given a source segment $x$, an MT model produces an initial hypothesis $y^{(0)}$ in the target language, where the ground truth target segment is $y^\ast$. We aim to produce improved translations $y^{(i)}$ for the iteration count $i = 1, 2, \cdots$, via an LLM-guided, retrieval-grounded post-edit step, dispensing with the costly MT model re-training.

Our design is intentionally modular, enforcing a strict separation of burden between the evaluator and post-editor: after the generation of $y^{(0)}$, an LLM evaluator first produces structured, MQM-style diagnoses, and post-editor LLMs sequentially apply targeted repairs based on the evaluator's error analysis and edit suggestion. This design avoids self-correction with automated feedback, which has been shown a challenging task particularly for complex reasoning tasks \citep{tie2025llmscorrectthemselvesbenchmark, pan2023automaticallycorrectinglargelanguage, madaan2023selfrefineiterativerefinementselffeedback}, and can also suppresses overcorrection, as the post-editors are not asked to broadly ``improve'' translations, which could degrade the performance, but asked to resolve specific, diagnosed issues. Another benefit of our design is that it gives flexibility to apply additional post-editing steps as needed and easily pair different models depending on use cases and upon newer model releases.

Our system orchestrates retrieval, evaluation, correction, and stopping, imposed via prompts and hard constraints. Specifically, it:

\begin{enumerate}[itemsep=0mm]
    \item retrieves $k$ semantically similar segments from the domain knowledge base (KB) $\mathcal D$;
    
    \item evaluates once to obtain the MQM-style report $r$, and stops if no major or minor issues are reported;
    
    \item otherwise applies additional post-editing steps as needed.
\end{enumerate}

In subsequent subsections, we elaborate on the responsibilities and behaviors of the evaluator and post-editors in a precise way.

\subsection{Edit Contracts} \label{subsec:edit-contracts}

In our design of the system, we identify that the separation of evaluator and post-editor responsibilities may give rise to complications, as delegating the post-editing duty to a separate model may bring about undesired consequences. To mitigate such concern, we introduce \textbf{edit contracts} as a section of the prompt, bridging the gap between the evaluator's requests and post-editor's requirements.

As part of the edit contract, the evaluator provides an explicit guidance on suggested fixes on each issue type, and post-editors are required to act based on the specific terms. The expectation is that the post-editors are able to focus on generating edits that are attributed to evaluator issues and/or retrieved evidence, hence effectively decreasing the likelihood of overcorrection. We provide our form of edit contract in Appendix~\ref{sec:evaluator-and-post-editor-prompt}.

\subsection{Structured MQM Evaluator} \label{subsec:structured-mqm-evaluator}

An evaluator $E$ is an LLM responsible for analyzing $y^{(0)}$ and generating $r$ in a reference-free manner. The prompt given to $E$ is comprised of the static portion $p_E$ (minimally including the task description, MQM rubric, output schema, and evaluator-side edit contract) and dynamic information that differs per query (such as $x$, $y^{(0)}$, and $k$ semantically most similar exemplars $\{s_i\}_{i=1}^k$ retrieved from the KB $\mathcal D$ populated with high-quality source-target segment pairs for the purpose of providing domain-specific knowledge to the evaluator). Formally, we have:
\begin{align*}
    r = \{e_j\}_{j=1}^{n} = E(p_E, x, y^{(0)}, \{s_i\}_{i=1}^k),
\end{align*}

where $e_j$ denotes an analysis of the diagnosed issue, generated based on the required schema, which contains fields such as \texttt{issue\_id}, \texttt{span}, \texttt{category}, \texttt{severity} $\in \{ \texttt{major}, \texttt{minor}, \texttt{neutral} \}$, and a short \texttt{description}, optionally with fields such as \texttt{suggested\_fix} and \texttt{evidence\_ids} linking to retrieved items.

In our system, we run the evaluator only once per query, producing $r$ for the initial hypothesis $y^{(0)}$.

\subsection{Post-Editors} \label{subsec:post-editors}

If $r$ contains any issues, we send the evaluator output and $y^{(0)}$ to the post-editors. Each post-editor outputs a new hypothesis $y^{(i)}$ and an edit log $\ell^{(i)}$ describing each change made, which should reference one or more \texttt{issue\_id}'s. We explicitly instruct the post-editors by including a statement within its edit contract section of the static prompt $p_C$ that they should prioritize minimal edits, avoiding paraphrases of unaffected spans. To formalize, a post-editor $C_i$ is an LLM that receives a 4-tuple $(p_C, x , r, \{y^{(j)}\}_{j=0}^{i-1})$ as part of its prompt, where $p_C$ denotes the static task description for the post-editor and $\{y^{(j)}\}$ a compact history of hypotheses generated up to this point. The post-editor prompt is intentionally concise and constrained to diagnosis-conditioned repairs (see Appendix~\ref{sec:evaluator-and-post-editor-prompt}), while the inclusion of history enables later post-editors to reference earlier edits and decisions made by the previous set of post-editors.

The modularized aspect of the system also allows for applying the post-editing step more than once, to a sequence of post-editors $\{C_1, C_2, \cdots \}$. For $C_i$ where $i \geq 2$, it first assesses whether $y^{(i-1)}$ resolves all errors. If it judges so, it immediately exits; otherwise, it generates a 2-tuple $(y^{(i)}, \ell^{(i)})$, and potentially continues iterating afterwards.

\section{Experiments} \label{sec:experiments}

\subsection{Experimental Setup} \label{subsec:experimental-setup}

\textbf{Data.}  We evaluate on internal Translation Memory representing the e-commerce domain, with 5k random samples as the held-out stratified test set and $\mathcal D$ built from non-test segments. Full dataset details are available in Appendix \ref{sec:model-dataset-details}.

\vspace{5pt}
\textbf{Languages.}  We focus on translating English (\textsc{EN}) segments to six languages: German (\textsc{DE}), Italian (\textsc{IT}), Spanish (\textsc{ES}), French (\textsc{FR}), Chinese (\textsc{CN}), and Japanese (\textsc{JP}).

\vspace{5pt}
\textbf{Models.}  We use Anthropic's Claude 3.5 Sonnet and Claude Opus 4.5 as our MT model generating $y^{(0)}$ and our evaluator, respectively. For post-editors, we experiment with a diverse set of LLMs available on Amazon Bedrock, spanning multiple providers and sizes. For a complete list of LLMs, refer to Appendix~\ref{sec:model-dataset-details}. We use Amazon OpenSearch Service for KB construction and retrieval.

\vspace{5pt}
\textbf{Evaluation.}  We report the quality of generated translations using Unbabel's reference-based evaluation metric \textsc{Comet}-22 \citep{rei-etal-2022-comet} and reference-free evaluation metric \textsc{CometKiwi} \citep{rei-etal-2022-cometkiwi}, together of which can provide a holistic picture of how our system performs.


\subsection{Experimental Design} \label{subsec:experimental-design}

The following steps delineate the end-to-end process of our experiments:
\begin{enumerate}
    \item The MT model generates $y^{(0)}$, which is passed to the MQM evaluator. The evaluator prompt, illustrated in Appendix \ref{sec:evaluator-and-post-editor-prompt}, is built on GEMBA-MQM but further enhanced by incorporating RAG, including a descriptive rubric, and specifying the edit contract (Section \ref{subsec:edit-contracts}).

    \item In the first round of post-editing, each post-editor in our LLM universe analyzes $y^{(0)}$, referencing the evaluator output, and generates $y^{(1)}$. This process yields the post-editor $C_1^*$ that makes the biggest improvement of post-edit quality on average. We note that this is a fair comparison, for $y^{(0)}$ and the evaluator output are shared across all post-editors.

    \item In the second round of post-editing, we pass the output of $C_1^*$ to the same set of LLMs, where each post-editor can either conclude all errors have been addressed and therefore exit, or flag the presence of errors and post-edit $y^{(1)}$ to generate $y^{(2)}$.

    \item We can optionally iterate this process by feeding the output of $C_i^*$ and evaluator output to $C_{i+1}$'s.
\end{enumerate}

\subsection{Experiment Variations} \label{subsec:experiment-variations}
To assess the effectiveness of our prompt and utilizing separate post-editors in the first place, we propose the following set of additional experiments differing in terms of the evaluator prompts:
\newpage
\begin{enumerate}[label=\textbf{P\arabic*.}]
    \item GEMBA-MQM prompt \citep{kocmi2023gembamqmdetectingtranslationquality};
    \item GEMBA-MQM prompt $+$ descriptive rubric $+$ edit contract;
    \item GEMBA-MQM prompt $+$ descriptive rubric $+$ edit contract $+$ RAG.
\end{enumerate}

Furthermore, to investigate the effectiveness of separating evaluators and post-editors, we propose running an additional experiment that combines evaluators and post-editors (\textit{i.e.}, having Claude Opus 4.5 analyze the errors contained within $y^{(0)}$ and also generate post-edits based on them), utilizing the best prompting strategy among \textbf{P1} through \textbf{P3} as empirically validated:

\begin{enumerate}[label=\textbf{P\arabic*.},start=4]
    \item Empirically the best prompting strategy on average from \textbf{P1} to \textbf{P3} $+$ post-edit instructions.
\end{enumerate}

\section{Results and Analysis} \label{sec:results-and-analysis}

In this section, we mainly focus on the results for \textsc{EN} $\to$ \textsc{DE} and \textsc{EN} $\to$ \textsc{CN}. However, our findings generalize across all six languages, demonstrating the effectiveness of our proposed approach. The full results can be found in Appendix~\ref{sec:all-results}.

\begin{table*}[t]
\centering
\begin{adjustbox}{width=\textwidth}
\begin{tabular}{c c c cc cc}
\toprule
\multirow{2}{*}{\textbf{Method}} &
\multirow{2}{*}{\textbf{Evaluator Prompt}} &
\multirow{2}{*}{\textbf{Post-editor}} &
\multicolumn{2}{c}{\textsc{EN} $\to$ \textsc{DE}} &
\multicolumn{2}{c}{\textsc{EN} $\to$ \textsc{CN}} \\
\cmidrule(lr){4-5}\cmidrule(lr){6-7}
& & &
\textsc{Comet} & \textsc{CometKiwi} &
\textsc{Comet} & \textsc{CometKiwi} \\
\midrule
\textit{No post-edit} & -- & -- &
86.39 & 82.66 &
87.76 & 81.60 \\ \midrule
\textit{One-stage} & \textbf{P3} & Claude Opus 4.5 &
89.38 {\color{Green} \footnotesize{(+2.99)}} & 82.84 {\color{Green} \footnotesize{(+0.18)}} &
88.70 {\color{Green} \footnotesize{(+0.94)}} & 79.57 {\color{red} \footnotesize{(-2.03)}} \\
\midrule
\textit{Two-stage} & \textbf{P1} & Claude Sonnet 4.5 &
86.47 {\color{Green} \footnotesize{(+0.08)}} & 82.65 {\color{red} \footnotesize{(-0.01)}} &
87.97 {\color{Green} \footnotesize{(+0.21)}} & 81.70 {\color{Green} \footnotesize{(+0.10)}} \\
\textit{Two-stage} & \textbf{P2} & Claude Sonnet 4.5 &
86.57 {\color{Green} \footnotesize{(+0.18)}} & 82.66 {\color{gray} \footnotesize{(+0.00)}} &
88.86 {\color{Green} \footnotesize{(+1.10)}} & 81.67 {\color{Green} \footnotesize{(+0.07)}} \\
\textit{Two-stage} & \textbf{P3} & Claude Sonnet 4.5 &
\textbf{89.93} {\color{Green} \footnotesize{(+3.54)}} & \textbf{84.50} {\color{Green} \footnotesize{(+1.84)}} &
\textbf{90.80} {\color{Green} \footnotesize{(+3.04)}} & \textbf{83.19} {\color{Green} \footnotesize{(+1.59)}} \\
\bottomrule
\end{tabular}
\end{adjustbox}
\caption{\textsc{Comet}-22 and \textsc{CometKiwi} scores of $y^{(1)}$ using one-stage and two-stage post-edit method, the latter with three prompt varieties; \textbf{P1} is GEMBA-MQM prompt, \textbf{P2} is MQM $+$ rubric $+$ edit contract, and \textbf{P3} is MQM $+$ rubric $+$ edit contract $+$ RAG. In this table, we have selected \textsc{EN} $\to$ \textsc{DE} and \textsc{EN} $\to$ \textsc{CN} for display. For complete results across all six target languages, refer to Appendix~\ref{sec:all-results}.}
\label{tab:main-de-cn}
\end{table*}

\subsection{Which Prompt Works the Best?}
The primary question concerns how different evaluator prompting strategies affect downstream post-edit quality. We compare three variants (Section~\ref{subsec:experiment-variations}): \textbf{P1} (GEMBA-MQM), \textbf{P2} (MQM $+$ descriptive rubric $+$ edit contract), and \textbf{P3} (\textbf{P2} + RAG with domain exemplars). 

Table~\ref{tab:main-de-cn} summarize \textsc{Comet}-22 and \textsc{CometKiwi} scores for $y^{(1)}$. Compared to \textbf{P1}, \textbf{P2} yields small but consistent improvements, indicating that descriptive rubrics and structured edit contracts improve the utility of evaluator outputs. We observe the largest gains under \textbf{P3}, where the retrieval from $\mathcal D$ further improves both metrics (notably for \textsc{EN} $\to$ \textsc{DE}). This suggests that domain exemplars can benefit the generated post-edits by reducing ambiguity in terminology and style, enabling more precise diagnoses and better-aligned edits. Importantly, these gains are consistently observed under a fixed post-editor LLM (Claude Sonnet 4.5), confirming that the evaluator design directly influences repair quality. Overall, structured grounding and constrained edit guidance outperform unconstrained refinement, improving post-edit quality while reducing unnecessary rewrites.

\subsection{Is a separate post-edit process effective?} \label{subsec:two-stage-vs-one-stage}

Table~\ref{tab:5-1} (and the full results in Table~\ref{tab:full} in the Appendix~\ref{sec:all-results}) shows that evaluator-guided post-editing helps, as measured by \textsc{Comet}-22 score improvements compared against the no post-edit baseline, and that the two-stage design is generally more effective than a one-stage ``judge-and-refine'' approach: 4 out of 7 models yield a higher \textsc{Comet} score for \textsc{EN} $\to$ \textsc{DE}, while 6 out of 7 models yield a higher \textsc{Comet}-22 score for \textsc{EN} $\to$ \textsc{CN}. Based on the substantial gains obtained by the two-stage method for \textsc{DE} and \textsc{CN}, we conclude that the two-stage post-edit process with our prompt design (Section~\ref{subsec:experiment-variations} and Appendix~\ref{sec:evaluator-and-post-editor-prompt}) systematically translates into measurable quality improvements without re-training the underlying model. 

Overall, these results support our central claim that there is a clear benefit derived from the separation of evaluators and post-editors, reducing unconstrained re-writing and better preserving already correct spans, thereby yielding better outputs overall while maintaining all the benefits that one-stage post-edit framework offers.

\begin{table*}[ht]
\centering

\begin{adjustbox}{width=0.9\textwidth}
\begin{tabular}{c c c c c c c c c c }
\toprule
\multirow{2}{*}{\textbf{Method}} & \multirow{2}{*}{\textbf{Model}} & \multicolumn{2}{c}{\textsc{EN} $\to$ \textsc{DE}} & \multicolumn{2}{c}{\textsc{EN} $\to$ \textsc{CN}} \\ \cmidrule(lr){3-4} \cmidrule(lr){5-6} \cmidrule(lr){7-8}
& & \textbf{\textsc{Comet}} & \textbf{\textsc{CometKiwi}} & \textbf{\textsc{Comet}} & \textbf{\textsc{CometKiwi}} \\ \midrule
\textit{No post-edit} & -- & 86.39 & 82.66 & 87.76 & 81.60 \\ \midrule
\textit{One-stage} & Claude Opus 4.5 & 89.38 {\color{Green} \footnotesize{(+2.99)}} & -- & 88.70 {\color{Green} \footnotesize{(+0.94)}} & -- \\ \midrule
\multirow{7}{*}{\textit{Two-stage}} 
& Gemma 3 4B & 88.29  {\color{Green} \footnotesize{(+1.90)}} 
& 82.43 {\color{Red} \footnotesize{(-0.23)}}
& \textbf{91.08} {\color{Green} \footnotesize{(+3.32)}}
& 79.82 {\color{Red} \footnotesize{(-1.78)}}\\
& Gemma 3 27B 
& 89.62 {\color{Green} \footnotesize{(+3.23)}} & 83.03 {\color{Green} \footnotesize{(+0.37)}} 
& 90.83 {\color{Green} \footnotesize{(+3.07)}}
& 81.92 {\color{Green} \footnotesize{(+0.32)}}\\
& Claude 3.5 Haiku 
& 89.53 {\color{Green} \footnotesize{(+3.14)}} 
& 82.69 {\color{Green} \footnotesize{(+0.03)}} 
& 90.20 {\color{Green} \footnotesize{(+2.44)}} 
& 82.51 {\color{Green} \footnotesize{(+0.91)}} 
\\
& Claude Sonnet 4 
& 89.85 {\color{Green} \footnotesize{(+3.46)}}  & 83.45 {\color{Green} \footnotesize{(+0.79)}} 
&  90.56 {\color{Green} \footnotesize{(+2.80)}} 
&  82.94 {\color{Green} \footnotesize{(+1.34)}} \\
& Claude Sonnet 4.5 
& \textbf{89.93} {\color{Green} \footnotesize{(+3.54)}} &\textbf{84.50} {\color{Green} \footnotesize{(+1.84)}} 
&  90.80 {\color{Green} \footnotesize{(+3.04)}} 
&  \textbf{83.19} {\color{Green} \footnotesize{(+1.59)}} \\
& GPT OSS 20B & 89.10 {\color{Green} \footnotesize{(+2.71)}} & 82.67 {\color{Green} \footnotesize{(+0.01)}} 
& 87.98 {\color{Green} \footnotesize{(+0.22)}} 
& 79.06 {\color{Red} \footnotesize{(-2.54)}} \\
& GPT OSS 120B & 89.18 {\color{Green} \footnotesize{(+2.79)}} & 82.76 {\color{Green} \footnotesize{(+0.10)}}
& 89.63 {\color{Green} \footnotesize{(+1.87)}} 
& 79.52 {\color{Red} \footnotesize{(-2.08)}} \\
\bottomrule
\end{tabular}

\end{adjustbox}
\caption{\textsc{Comet}-22 and \textsc{CometKiwi} scores of $y^{(0)}$ and $y^{(1)}$ across all considered LLMs, using \textbf{P3} as the evaluator prompting strategy}
\label{tab:5-1}
\end{table*}


\subsection{Which LLM generates the best post-edits?} 

Our LLM universe spans three model providers: Google, Anthropic, and OpenAI. From each model provider, we have carefully selected models on both ends of the spectrum in terms of model sizes, as measured by the number of parameters. This allows us to analyze how post-edit qualities differ across model providers and sizes.

From Table~\ref{tab:5-1}, we find that the Claude models show the strongest post-edit capability for \textsc{EN} $\to$ \textsc{DE} language direction, with Claude 3.5 Haiku showing higher \textsc{Comet}-22 scores than both GPT OSS models and similar scores compared to Gemma 3 27B model. \textsc{CometKiwi} scores also support a similar conclusion. 

For \textsc{EN} $\to$ \textsc{CN}, Gemma 3 is the overall winner, yielding higher \textsc{Comet}-22 scores compared to Claude and GPT OSS models. Interestingly, while Gemma 3 4B shows the smallest \textsc{Comet} score gain among all considered LLMs for \textsc{EN} $\to$ \textsc{DE} and even lower \textsc{CometKiwi} score than that of $y^{(0)}$, we find that it achieves the highest average \textsc{Comet} score for \textsc{EN} $\to$ \textsc{CN}. Larger models tend to show better post-edit qualities, with one notable exception being Gemma 3 4B for \textsc{EN} $\to$ \textsc{CN}. This observation suggests that there is not a single best model, making it necessary to experiment with a large universe of models with the data representative of the population prior to selecting the model for a specific language direction and domain.

\subsection{How often does $C_2$ think $y^{(1)}$ still contains errors?}

To quantify how much benefit a second post-editing round could yield (that is, $E$, followed by $C_1$, followed by $C_2$; refer to Section~\ref{subsec:post-editors}), we measure how often $C_2$ flags the first-round output $y^{(1)}$ (generated by Claude Sonnet 4.5) as still containing errors. Table~\ref{tab:c2-de} reports results for EN $\to$ DE.

\begin{table}[H]
\centering
\begin{adjustbox}{width=0.43\textwidth}
\begin{tabular}{l r r}
\toprule
\textbf{$C_2$ Model} & \multicolumn{1}{c}{\textbf{\# Flagged}} & \textbf{Rate (\%)} \\
\midrule
Claude 3.5 Haiku & 205 / 4031 & 5.1 \\
Claude Sonnet 4 & 875 / 4031 & 21.7 \\
Claude Sonnet 4.5 & 1282 / 4031 & 31.8 \\
Gemma 3 4B & 328 / 4031 & 8.1 \\
Gemma 3 27B & 541 / 4031 & 13.4 \\
GPT OSS 20B & 278 / 4031 & 6.9 \\
GPT OSS 120B & 207 / 4031 & 5.1 \\
\bottomrule
\end{tabular}
\end{adjustbox}
\caption{Continuation rate of $C_2$ after first-round correction ($y^{(1)}$) for EN $\to$ DE.}
\label{tab:c2-de}
\end{table}

Across 4,031 segments, most models judge the majority of first-round outputs as sufficiently corrected. For several models (Claude 3.5 Haiku, GPT OSS 120B), only around 5\% of segments are flagged for further editing, suggesting that a single targeted repair pass effectively resolves most diagnosed issues. In contrast, stronger Claude model variants tend to apply stricter criteria (Claude Sonnet 4 flags 21.7\% of segments, and Claude Sonnet 4.5 flags 31.8\%), indicating that higher-capacity reasoning models are more likely to detect residual minor or stylistic issues even after targeted post-edits.

Overall, these results suggest that one round of post-edit process is likely sufficient for the majority of segments, while additional rounds primarily serve to address stricter quality thresholds rather than widespread unresolved major errors.

\subsection{Are second round post-edits beneficial?}

We next evaluate whether applying post-editing with $C_2$ yields further quality gains over the first-round output $y^{(1)}$. Table~\ref{tab:c2-benefit-de} reports EN $\to$ DE results using the best-performing prompt configuration (\textbf{P3}). Claude Sonnet 4.5 serves as $C_1$, and we consider all other models in our universe as $C_2$.

\begin{table}[ht]
\centering
\begin{adjustbox}{width=0.33\textwidth}
\begin{tabular}{l c }
\toprule
\textbf{$C_2$ Model} & \textbf{$y^{(2)}$ \textsc{Comet}} \\
\midrule
Gemma 4B & 89.18 {\color{Red} \footnotesize{(-0.75)}} \\
Gemma 27B & 89.93 {\color{gray} \footnotesize{(+0.00)}} \\
Claude 3.5 Haiku & 89.63 {\color{Red} \footnotesize{(-0.30)}} \\
Claude Sonnet 4 & 89.31 {\color{Red} \footnotesize{(-0.62)}} \\
Claude Sonnet 4.5 & 90.07 {\color{Green} \footnotesize{(+0.14)}} \\
GPT OSS 20B & 89.30 {\color{Red} \footnotesize{(-0.63)}} \\
GPT OSS 120B & 88.07 {\color{Red} \footnotesize{(-1.86)}} \\
\bottomrule
\end{tabular}
\end{adjustbox}
\caption{Impact of second-round post-editing ($C_2$) on EN $\to$ DE}
\label{tab:c2-benefit-de}
\end{table}

The results show that a second post-editing round generally degrades post-edit quality: in six out of seven cases, \textsc{Comet}-22 scores decrease, sometimes substantially (for instance, $-$1.86 for GPT-OSS-120B). Only when the same high-capacity model (Claude Sonnet 4.5) is used again as $C_2$ do we observe a marginal improvement ($+$0.14), which is small relative to the first-round gains.

These findings suggest that most systematic errors are already resolved during the first evaluator-guided post-edit process. Additional rounds tend to introduce paraphrastic drift or unnecessary edits that degrade adequacy or fluency. In practice, a single structured evaluation followed by one targeted post-editing step appears sufficient, and further iterations offer limited benefit while increasing regression risk.

It is also worth noting that there could be a model-specific behavior at play. Combining the analysis from Tables~\ref{tab:c2-de} and \ref{tab:c2-benefit-de}, we find that Claude Sonnet 4.5, which is the excelling post-editor from previous experiments, is also more likely to flag issues even in the potential absence of real errors that should be pointed out. One possible reason for this behavior is the stronger instruction-following abilities obtained by the model as reinforced during the alignment process.

\section{Conclusion and Future Work} \label{sec:conclusion-and-future-work}

In this work, we proposed a two-stage post-edit framework for MT in specialized domains. By separating a retrieval-augmented evaluator from a constrained post-editor governed by edit contracts, the two-stage design delivers consistent gains across domain translation tasks on two language directions. Results from a systematic study involving seven LLMs show that an MQM-style, evidence-grounded evaluation can improve post-editor's correction quality and that a one-round targeted post-edits are sufficient for improving the post-edit quality, with further iteration offering only limited benefits. Future work includes exploring confidence-aware post-editing to further optimize cost-quality tradeoffs, and extending the framework to additional domains and language pairs.


\bibliography{custom}

@inproceedings{burchardt-2013-multidimensional,
    title = "Multidimensional quality metrics: a flexible system for assessing translation quality",
    author = "Lommel, Arle Richard  and
      Burchardt, Aljoscha  and
      Uszkoreit, Hans",
    booktitle = "Proceedings of Translating and the Computer 35",
    month = nov # " 28-29",
    year = "2013",
    address = "London, UK",
    publisher = "Aslib",
    url = "https://aclanthology.org/2013.tc-1.6/"
}

@inproceedings{lommel-etal-2014-using,
    title = "Using a new analytic measure for the annotation and analysis of {MT} errors on real data",
    author = "Lommel, Arle  and
      Burchardt, Aljoscha  and
      Popovi{\'c}, Maja  and
      Harris, Kim  and
      Avramidis, Eleftherios  and
      Uszkoreit, Hans",
    editor = "Cettolo, Mauro  and
      Federico, Marcello  and
      Specia, Lucia  and
      Way, Andy",
    booktitle = "Proceedings of the 17th Annual Conference of the European Association for Machine Translation",
    month = jun # " 16-18",
    year = "2014",
    address = "Dubrovnik, Croatia",
    publisher = "European Association for Machine Translation",
    url = "https://aclanthology.org/2014.eamt-1.38/",
    pages = "165--172"
}

@article{freitag-etal-2021-experts,
    title = "Experts, Errors, and Context: A Large-Scale Study of Human Evaluation for Machine Translation",
    author = "Freitag, Markus  and
      Foster, George  and
      Grangier, David  and
      Ratnakar, Viresh  and
      Tan, Qijun  and
      Macherey, Wolfgang",
    editor = "Roark, Brian  and
      Nenkova, Ani",
    journal = "Transactions of the Association for Computational Linguistics",
    volume = "9",
    year = "2021",
    address = "Cambridge, MA",
    publisher = "MIT Press",
    url = "https://aclanthology.org/2021.tacl-1.87/",
    doi = "10.1162/tacl_a_00437",
    pages = "1460--1474"
}

@inproceedings{freitag-etal-2022-results,
    title = "Results of {WMT}22 Metrics Shared Task: Stop Using {BLEU} {--} Neural Metrics Are Better and More Robust",
    author = "Freitag, Markus  and
      Rei, Ricardo  and
      Mathur, Nitika  and
      Lo, Chi-kiu  and
      Stewart, Craig  and
      Avramidis, Eleftherios  and
      Kocmi, Tom  and
      Foster, George  and
      Lavie, Alon  and
      Martins, Andr{\'e} F. T.",
    editor = {Koehn, Philipp  and
      Barrault, Lo{\"i}c  and
      Bojar, Ond{\v{r}}ej  and
      Bougares, Fethi  and
      Chatterjee, Rajen  and
      Costa-juss{\`a}, Marta R.  and
      Federmann, Christian  and
      Fishel, Mark  and
      Fraser, Alexander  and
      Freitag, Markus  and
      Graham, Yvette  and
      Grundkiewicz, Roman  and
      Guzman, Paco  and
      Haddow, Barry  and
      Huck, Matthias  and
      Jimeno Yepes, Antonio  and
      Kocmi, Tom  and
      Martins, Andr{\'e}  and
      Morishita, Makoto  and
      Monz, Christof  and
      Nagata, Masaaki  and
      Nakazawa, Toshiaki  and
      Negri, Matteo  and
      N{\'e}v{\'e}ol, Aur{\'e}lie  and
      Neves, Mariana  and
      Popel, Martin  and
      Turchi, Marco  and
      Zampieri, Marcos},
    booktitle = "Proceedings of the Seventh Conference on Machine Translation (WMT)",
    month = dec,
    year = "2022",
    address = "Abu Dhabi, United Arab Emirates (Hybrid)",
    publisher = "Association for Computational Linguistics",
    url = "https://aclanthology.org/2022.wmt-1.2/",
    doi = "10.18653/v1/2022.wmt-1.2",
    pages = "46--68"
}

@misc{zheng2023judgingllmasajudgemtbenchchatbot,
      title={Judging LLM-as-a-Judge with MT-Bench and Chatbot Arena}, 
      author={Lianmin Zheng and Wei-Lin Chiang and Ying Sheng and Siyuan Zhuang and Zhanghao Wu and Yonghao Zhuang and Zi Lin and Zhuohan Li and Dacheng Li and Eric P. Xing and Hao Zhang and Joseph E. Gonzalez and Ion Stoica},
      year={2023},
      eprint={2306.05685},
      archivePrefix={arXiv},
      primaryClass={cs.CL},
      url={https://arxiv.org/abs/2306.05685}, 
}

@inproceedings{liu-etal-2023-g,
    title = "{G}-Eval: {NLG} Evaluation using Gpt-4 with Better Human Alignment",
    author = "Liu, Yang  and
      Iter, Dan  and
      Xu, Yichong  and
      Wang, Shuohang  and
      Xu, Ruochen  and
      Zhu, Chenguang",
    editor = "Bouamor, Houda  and
      Pino, Juan  and
      Bali, Kalika",
    booktitle = "Proceedings of the 2023 Conference on Empirical Methods in Natural Language Processing",
    month = dec,
    year = "2023",
    address = "Singapore",
    publisher = "Association for Computational Linguistics",
    url = "https://aclanthology.org/2023.emnlp-main.153/",
    doi = "10.18653/v1/2023.emnlp-main.153",
    pages = "2511--2522"
}

@inproceedings{chiang-lee-2023-closer,
    title = "A Closer Look into Using Large Language Models for Automatic Evaluation",
    author = "Chiang, Cheng-Han  and
      Lee, Hung-yi",
    editor = "Bouamor, Houda  and
      Pino, Juan  and
      Bali, Kalika",
    booktitle = "Findings of the Association for Computational Linguistics: EMNLP 2023",
    month = dec,
    year = "2023",
    address = "Singapore",
    publisher = "Association for Computational Linguistics",
    url = "https://aclanthology.org/2023.findings-emnlp.599/",
    doi = "10.18653/v1/2023.findings-emnlp.599",
    pages = "8928--8942"
}

@misc{verga2024replacingjudgesjuriesevaluating,
      title={Replacing Judges with Juries: Evaluating LLM Generations with a Panel of Diverse Models}, 
      author={Pat Verga and Sebastian Hofstatter and Sophia Althammer and Yixuan Su and Aleksandra Piktus and Arkady Arkhangorodsky and Minjie Xu and Naomi White and Patrick Lewis},
      year={2024},
      eprint={2404.18796},
      archivePrefix={arXiv},
      primaryClass={cs.CL},
      url={https://arxiv.org/abs/2404.18796}, 
}

@misc{kocmi2023gembamqmdetectingtranslationquality,
      title={GEMBA-MQM: Detecting Translation Quality Error Spans with GPT-4}, 
      author={Tom Kocmi and Christian Federmann},
      year={2023},
      eprint={2310.13988},
      archivePrefix={arXiv},
      primaryClass={cs.CL},
      url={https://arxiv.org/abs/2310.13988}, 
}

@inproceedings{kim-2025-rubric,
    title = "{RUBRIC}-{MQM} : Span-Level {LLM}-as-judge in Machine Translation For High-End Models",
    author = "Kim, Ahrii",
    editor = "Rehm, Georg  and
      Li, Yunyao",
    booktitle = "Proceedings of the 63rd Annual Meeting of the Association for Computational Linguistics (Volume 6: Industry Track)",
    month = jul,
    year = "2025",
    address = "Vienna, Austria",
    publisher = "Association for Computational Linguistics",
    url = "https://aclanthology.org/2025.acl-industry.12/",
    doi = "10.18653/v1/2025.acl-industry.12",
    pages = "147--165",
    ISBN = "979-8-89176-288-6"
}

@inproceedings{chen-etal-2024-iterative,
    title = "Iterative Translation Refinement with Large Language Models",
    author = "Chen, Pinzhen  and
      Guo, Zhicheng  and
      Haddow, Barry  and
      Heafield, Kenneth",
    editor = "Scarton, Carolina  and
      Prescott, Charlotte  and
      Bayliss, Chris  and
      Oakley, Chris  and
      Wright, Joanna  and
      Wrigley, Stuart  and
      Song, Xingyi  and
      Gow-Smith, Edward  and
      Bawden, Rachel  and
      S{\'a}nchez-Cartagena, V{\'i}ctor M  and
      Cadwell, Patrick  and
      Lapshinova-Koltunski, Ekaterina  and
      Cabarr{\~a}o, Vera  and
      Chatzitheodorou, Konstantinos  and
      Nurminen, Mary  and
      Kanojia, Diptesh  and
      Moniz, Helena",
    booktitle = "Proceedings of the 25th Annual Conference of the European Association for Machine Translation (Volume 1)",
    month = jun,
    year = "2024",
    address = "Sheffield, UK",
    publisher = "European Association for Machine Translation (EAMT)",
    url = "https://aclanthology.org/2024.eamt-1.17/",
    pages = "181--190"
}

@inproceedings{ki-carpuat-2024-guiding,
    title = "Guiding Large Language Models to Post-Edit Machine Translation with Error Annotations",
    author = "Ki, Dayeon  and
      Carpuat, Marine",
    editor = "Duh, Kevin  and
      Gomez, Helena  and
      Bethard, Steven",
    booktitle = "Findings of the Association for Computational Linguistics: NAACL 2024",
    month = jun,
    year = "2024",
    address = "Mexico City, Mexico",
    publisher = "Association for Computational Linguistics",
    url = "https://aclanthology.org/2024.findings-naacl.265/",
    doi = "10.18653/v1/2024.findings-naacl.265",
    pages = "4253--4273"
}

@misc{lewis2021retrievalaugmentedgenerationknowledgeintensivenlp,
      title={Retrieval-Augmented Generation for Knowledge-Intensive NLP Tasks}, 
      author={Patrick Lewis and Ethan Perez and Aleksandra Piktus and Fabio Petroni and Vladimir Karpukhin and Naman Goyal and Heinrich Küttler and Mike Lewis and Wen-tau Yih and Tim Rocktäschel and Sebastian Riedel and Douwe Kiela},
      year={2021},
      eprint={2005.11401},
      archivePrefix={arXiv},
      primaryClass={cs.CL},
      url={https://arxiv.org/abs/2005.11401}, 
}

@inproceedings{zhang-etal-2018-guiding,
    title = "Guiding Neural Machine Translation with Retrieved Translation Pieces",
    author = "Zhang, Jingyi  and
      Utiyama, Masao  and
      Sumita, Eiichro  and
      Neubig, Graham  and
      Nakamura, Satoshi",
    editor = "Walker, Marilyn  and
      Ji, Heng  and
      Stent, Amanda",
    booktitle = "Proceedings of the 2018 Conference of the North {A}merican Chapter of the Association for Computational Linguistics: Human Language Technologies, Volume 1 (Long Papers)",
    month = jun,
    year = "2018",
    address = "New Orleans, Louisiana",
    publisher = "Association for Computational Linguistics",
    url = "https://aclanthology.org/N18-1120/",
    doi = "10.18653/v1/N18-1120",
    pages = "1325--1335"
}

@inproceedings{bulte-tezcan-2019-neural,
    title = "Neural Fuzzy Repair: Integrating Fuzzy Matches into Neural Machine Translation",
    author = "Bulte, Bram  and
      Tezcan, Arda",
    editor = "Korhonen, Anna  and
      Traum, David  and
      M{\`a}rquez, Llu{\'i}s",
    booktitle = "Proceedings of the 57th Annual Meeting of the Association for Computational Linguistics",
    month = jul,
    year = "2019",
    address = "Florence, Italy",
    publisher = "Association for Computational Linguistics",
    url = "https://aclanthology.org/P19-1175/",
    doi = "10.18653/v1/P19-1175",
    pages = "1800--1809"
}

@inproceedings{conia-etal-2024-towards,
    title = "Towards Cross-Cultural Machine Translation with Retrieval-Augmented Generation from Multilingual Knowledge Graphs",
    author = "Conia, Simone  and
      Lee, Daniel  and
      Li, Min  and
      Minhas, Umar Farooq  and
      Potdar, Saloni  and
      Li, Yunyao",
    editor = "Al-Onaizan, Yaser  and
      Bansal, Mohit  and
      Chen, Yun-Nung",
    booktitle = "Proceedings of the 2024 Conference on Empirical Methods in Natural Language Processing",
    month = nov,
    year = "2024",
    address = "Miami, Florida, USA",
    publisher = "Association for Computational Linguistics",
    url = "https://aclanthology.org/2024.emnlp-main.914/",
    doi = "10.18653/v1/2024.emnlp-main.914",
    pages = "16343--16360"
}

@misc{khandelwal2021nearestneighbormachinetranslation,
      title={Nearest Neighbor Machine Translation}, 
      author={Urvashi Khandelwal and Angela Fan and Dan Jurafsky and Luke Zettlemoyer and Mike Lewis},
      year={2021},
      eprint={2010.00710},
      archivePrefix={arXiv},
      primaryClass={cs.CL},
      url={https://arxiv.org/abs/2010.00710}, 
}

@misc{wang2025retrievalaugmentedmachinetranslationunstructured,
      title={Retrieval-Augmented Machine Translation with Unstructured Knowledge}, 
      author={Jiaan Wang and Fandong Meng and Yingxue Zhang and Jie Zhou},
      year={2025},
      eprint={2412.04342},
      archivePrefix={arXiv},
      primaryClass={cs.CL},
      url={https://arxiv.org/abs/2412.04342}, 
}

@inproceedings{simard-etal-2007-statistical,
    title = "Statistical Phrase-Based Post-Editing",
    author = "Simard, Michel  and
      Goutte, Cyril  and
      Isabelle, Pierre",
    editor = "Sidner, Candace  and
      Schultz, Tanja  and
      Stone, Matthew  and
      Zhai, ChengXiang",
    booktitle = "Human Language Technologies 2007: The Conference of the North {A}merican Chapter of the Association for Computational Linguistics; Proceedings of the Main Conference",
    month = apr,
    year = "2007",
    address = "Rochester, New York",
    publisher = "Association for Computational Linguistics",
    url = "https://aclanthology.org/N07-1064/",
    pages = "508--515"
}

@inproceedings{chatterjee-etal-2018-findings,
    title = "Findings of the {WMT} 2018 Shared Task on Automatic Post-Editing",
    author = "Chatterjee, Rajen  and
      Negri, Matteo  and
      Rubino, Raphael  and
      Turchi, Marco",
    editor = "Bojar, Ond{\v{r}}ej  and
      Chatterjee, Rajen  and
      Federmann, Christian  and
      Fishel, Mark  and
      Graham, Yvette  and
      Haddow, Barry  and
      Huck, Matthias  and
      Yepes, Antonio Jimeno  and
      Koehn, Philipp  and
      Monz, Christof  and
      Negri, Matteo  and
      N{\'e}v{\'e}ol, Aur{\'e}lie  and
      Neves, Mariana  and
      Post, Matt  and
      Specia, Lucia  and
      Turchi, Marco  and
      Verspoor, Karin",
    booktitle = "Proceedings of the Third Conference on Machine Translation: Shared Task Papers",
    month = oct,
    year = "2018",
    address = "Belgium, Brussels",
    publisher = "Association for Computational Linguistics",
    url = "https://aclanthology.org/W18-6452/",
    doi = "10.18653/v1/W18-6452",
    pages = "710--725"
}

@misc{tie2025llmscorrectthemselvesbenchmark,
      title={Can LLMs Correct Themselves? A Benchmark of Self-Correction in LLMs}, 
      author={Guiyao Tie and Zenghui Yuan and Zeli Zhao and Chaoran Hu and Tianhe Gu and Ruihang Zhang and Sizhe Zhang and Junran Wu and Xiaoyue Tu and Ming Jin and Qingsong Wen and Lixing Chen and Pan Zhou and Lichao Sun},
      year={2025},
      eprint={2510.16062},
      archivePrefix={arXiv},
      primaryClass={cs.CL},
      url={https://arxiv.org/abs/2510.16062}, 
}

@misc{pan2023automaticallycorrectinglargelanguage,
      title={Automatically Correcting Large Language Models: Surveying the landscape of diverse self-correction strategies}, 
      author={Liangming Pan and Michael Saxon and Wenda Xu and Deepak Nathani and Xinyi Wang and William Yang Wang},
      year={2023},
      eprint={2308.03188},
      archivePrefix={arXiv},
      primaryClass={cs.CL},
      url={https://arxiv.org/abs/2308.03188}, 
}

@misc{madaan2023selfrefineiterativerefinementselffeedback,
      title={Self-Refine: Iterative Refinement with Self-Feedback}, 
      author={Aman Madaan and Niket Tandon and Prakhar Gupta and Skyler Hallinan and Luyu Gao and Sarah Wiegreffe and Uri Alon and Nouha Dziri and Shrimai Prabhumoye and Yiming Yang and Shashank Gupta and Bodhisattwa Prasad Majumder and Katherine Hermann and Sean Welleck and Amir Yazdanbakhsh and Peter Clark},
      year={2023},
      eprint={2303.17651},
      archivePrefix={arXiv},
      primaryClass={cs.CL},
      url={https://arxiv.org/abs/2303.17651}, 
}

@misc{lu2024mqmapehighqualityerrorannotation,
      title={MQM-APE: Toward High-Quality Error Annotation Predictors with Automatic Post-Editing in LLM Translation Evaluators}, 
      author={Qingyu Lu and Liang Ding and Kanjian Zhang and Jinxia Zhang and Dacheng Tao},
      year={2024},
      eprint={2409.14335},
      archivePrefix={arXiv},
      primaryClass={cs.CL},
      url={https://arxiv.org/abs/2409.14335}, 
}

@inproceedings{rei-etal-2022-cometkiwi,
    title = "{C}omet{K}iwi: {IST}-Unbabel 2022 Submission for the Quality Estimation Shared Task",
    author = "Rei, Ricardo  and
      Treviso, Marcos  and
      Guerreiro, Nuno M.  and
      Zerva, Chrysoula  and
      Farinha, Ana C  and
      Maroti, Christine  and
      C. de Souza, Jos{\'e} G.  and
      Glushkova, Taisiya  and
      Alves, Duarte  and
      Coheur, Luisa  and
      Lavie, Alon  and
      Martins, Andr{\'e} F. T.",
    editor = {Koehn, Philipp  and
      Barrault, Lo{\"i}c  and
      Bojar, Ond{\v{r}}ej  and
      Bougares, Fethi  and
      Chatterjee, Rajen  and
      Costa-juss{\`a}, Marta R.  and
      Federmann, Christian  and
      Fishel, Mark  and
      Fraser, Alexander  and
      Freitag, Markus  and
      Graham, Yvette  and
      Grundkiewicz, Roman  and
      Guzman, Paco  and
      Haddow, Barry  and
      Huck, Matthias  and
      Jimeno Yepes, Antonio  and
      Kocmi, Tom  and
      Martins, Andr{\'e}  and
      Morishita, Makoto  and
      Monz, Christof  and
      Nagata, Masaaki  and
      Nakazawa, Toshiaki  and
      Negri, Matteo  and
      N{\'e}v{\'e}ol, Aur{\'e}lie  and
      Neves, Mariana  and
      Popel, Martin  and
      Turchi, Marco  and
      Zampieri, Marcos},
    booktitle = "Proceedings of the Seventh Conference on Machine Translation (WMT)",
    month = dec,
    year = "2022",
    address = "Abu Dhabi, United Arab Emirates (Hybrid)",
    publisher = "Association for Computational Linguistics",
    url = "https://aclanthology.org/2022.wmt-1.60/",
    doi = "10.18653/v1/2022.wmt-1.60",
    pages = "634--645"
}

@inproceedings{rei-etal-2022-comet,
    title = "{COMET}-22: Unbabel-{IST} 2022 Submission for the Metrics Shared Task",
    author = "Rei, Ricardo  and
      C. de Souza, Jos{\'e} G.  and
      Alves, Duarte  and
      Zerva, Chrysoula  and
      Farinha, Ana C  and
      Glushkova, Taisiya  and
      Lavie, Alon  and
      Coheur, Luisa  and
      Martins, Andr{\'e} F. T.",
    editor = {Koehn, Philipp  and
      Barrault, Lo{\"i}c  and
      Bojar, Ond{\v{r}}ej  and
      Bougares, Fethi  and
      Chatterjee, Rajen  and
      Costa-juss{\`a}, Marta R.  and
      Federmann, Christian  and
      Fishel, Mark  and
      Fraser, Alexander  and
      Freitag, Markus  and
      Graham, Yvette  and
      Grundkiewicz, Roman  and
      Guzman, Paco  and
      Haddow, Barry  and
      Huck, Matthias  and
      Jimeno Yepes, Antonio  and
      Kocmi, Tom  and
      Martins, Andr{\'e}  and
      Morishita, Makoto  and
      Monz, Christof  and
      Nagata, Masaaki  and
      Nakazawa, Toshiaki  and
      Negri, Matteo  and
      N{\'e}v{\'e}ol, Aur{\'e}lie  and
      Neves, Mariana  and
      Popel, Martin  and
      Turchi, Marco  and
      Zampieri, Marcos},
    booktitle = "Proceedings of the Seventh Conference on Machine Translation (WMT)",
    month = dec,
    year = "2022",
    address = "Abu Dhabi, United Arab Emirates (Hybrid)",
    publisher = "Association for Computational Linguistics",
    url = "https://aclanthology.org/2022.wmt-1.52/",
    doi = "10.18653/v1/2022.wmt-1.52",
    pages = "578--585"
}

@misc{koehn2017challengesneuralmachinetranslation,
      title={Six Challenges for Neural Machine Translation}, 
      author={Philipp Koehn and Rebecca Knowles},
      year={2017},
      eprint={1706.03872},
      archivePrefix={arXiv},
      primaryClass={cs.CL},
      url={https://arxiv.org/abs/1706.03872}, 
}

@misc{jiao2023chatgptgoodtranslatoryes,
      title={Is ChatGPT A Good Translator? Yes With GPT-4 As The Engine}, 
      author={Wenxiang Jiao and Wenxuan Wang and Jen-tse Huang and Xing Wang and Shuming Shi and Zhaopeng Tu},
      year={2023},
      eprint={2301.08745},
      archivePrefix={arXiv},
      primaryClass={cs.CL},
      url={https://arxiv.org/abs/2301.08745}, 
}

@inproceedings{alves-etal-2023-steering,
    title = "Steering Large Language Models for Machine Translation with Finetuning and In-Context Learning",
    author = "Alves, Duarte  and
      Guerreiro, Nuno  and
      Alves, Jo{\~a}o  and
      Pombal, Jos{\'e}  and
      Rei, Ricardo  and
      de Souza, Jos{\'e}  and
      Colombo, Pierre  and
      Martins, Andre",
    editor = "Bouamor, Houda  and
      Pino, Juan  and
      Bali, Kalika",
    booktitle = "Findings of the Association for Computational Linguistics: EMNLP 2023",
    month = dec,
    year = "2023",
    address = "Singapore",
    publisher = "Association for Computational Linguistics",
    url = "https://aclanthology.org/2023.findings-emnlp.744/",
    doi = "10.18653/v1/2023.findings-emnlp.744",
    pages = "11127--11148"
}

@inproceedings{wang-etal-2023-document-level,
    title = "Document-Level Machine Translation with Large Language Models",
    author = "Wang, Longyue  and
      Lyu, Chenyang  and
      Ji, Tianbo  and
      Zhang, Zhirui  and
      Yu, Dian  and
      Shi, Shuming  and
      Tu, Zhaopeng",
    editor = "Bouamor, Houda  and
      Pino, Juan  and
      Bali, Kalika",
    booktitle = "Proceedings of the 2023 Conference on Empirical Methods in Natural Language Processing",
    month = dec,
    year = "2023",
    address = "Singapore",
    publisher = "Association for Computational Linguistics",
    url = "https://aclanthology.org/2023.emnlp-main.1036/",
    doi = "10.18653/v1/2023.emnlp-main.1036",
    pages = "16646--16661"
}

@inproceedings{mathur-etal-2020-tangled,
    title = "Tangled up in {BLEU}: Reevaluating the Evaluation of Automatic Machine Translation Evaluation Metrics",
    author = "Mathur, Nitika  and
      Baldwin, Timothy  and
      Cohn, Trevor",
    editor = "Jurafsky, Dan  and
      Chai, Joyce  and
      Schluter, Natalie  and
      Tetreault, Joel",
    booktitle = "Proceedings of the 58th Annual Meeting of the Association for Computational Linguistics",
    month = jul,
    year = "2020",
    address = "Online",
    publisher = "Association for Computational Linguistics",
    url = "https://aclanthology.org/2020.acl-main.448/",
    doi = "10.18653/v1/2020.acl-main.448",
    pages = "4984--4997"
}

@inproceedings{rei-etal-2020-comet,
    title = "{COMET}: A Neural Framework for {MT} Evaluation",
    author = "Rei, Ricardo  and
      Stewart, Craig  and
      Farinha, Ana C  and
      Lavie, Alon",
    editor = "Webber, Bonnie  and
      Cohn, Trevor  and
      He, Yulan  and
      Liu, Yang",
    booktitle = "Proceedings of the 2020 Conference on Empirical Methods in Natural Language Processing (EMNLP)",
    month = nov,
    year = "2020",
    address = "Online",
    publisher = "Association for Computational Linguistics",
    url = "https://aclanthology.org/2020.emnlp-main.213/",
    doi = "10.18653/v1/2020.emnlp-main.213",
    pages = "2685--2702"
}

@misc{huang2024largelanguagemodelsselfcorrect,
      title={Large Language Models Cannot Self-Correct Reasoning Yet}, 
      author={Jie Huang and Xinyun Chen and Swaroop Mishra and Huaixiu Steven Zheng and Adams Wei Yu and Xinying Song and Denny Zhou},
      year={2024},
      eprint={2310.01798},
      archivePrefix={arXiv},
      primaryClass={cs.CL},
      url={https://arxiv.org/abs/2310.01798}, 
}

@misc{wu2025translateagainsimpleexperiments,
      title={Please Translate Again: Two Simple Experiments on Whether Human-Like Reasoning Helps Translation}, 
      author={Di Wu and Seth Aycock and Christof Monz},
      year={2025},
      eprint={2506.04521},
      archivePrefix={arXiv},
      primaryClass={cs.CL},
      url={https://arxiv.org/abs/2506.04521}, 
}

@misc{briakou2024translatingstepbystepdecomposingtranslation,
      title={Translating Step-by-Step: Decomposing the Translation Process for Improved Translation Quality of Long-Form Texts}, 
      author={Eleftheria Briakou and Jiaming Luo and Colin Cherry and Markus Freitag},
      year={2024},
      eprint={2409.06790},
      archivePrefix={arXiv},
      primaryClass={cs.CL},
      url={https://arxiv.org/abs/2409.06790}, 
}

\clearpage
\onecolumn
\appendix

\section*{Appendix}

\section{Complete Results} \label{sec:all-results}

\begin{table*}[ht]
\centering

\begin{adjustbox}{width=\textwidth}
\begin{tabular}{c c c c c c c c}
\toprule
\multirow{2}{*}{\textbf{Method}} & \multirow{2}{*}{\textbf{Model}} & \multicolumn{2}{c}{\textsc{EN} $\to$ \textsc{DE}} & \multicolumn{2}{c}{\textsc{EN} $\to$ \textsc{CN}} & \multicolumn{2}{c}{\textsc{EN} $\to$ \textsc{FR}} \\ \cmidrule(lr){3-4} \cmidrule(lr){5-6} \cmidrule(lr){7-8}
& & \textbf{\textsc{Comet}} & \textbf{\textsc{CometKiwi}} & \textbf{\textsc{Comet}} & \textbf{\textsc{CometKiwi}} & \textbf{\textsc{Comet}} & \textbf{\textsc{CometKiwi}} \\ \midrule
\textit{No post-edit} & -- & 86.39 & 82.66 & 87.76 & 81.60 & 87.19 & 82.65 \\ \midrule
\multirow{1}{*}{\textit{One-stage}} & Claude Opus 4.5 & 89.38 {\color{Green} \footnotesize{(+2.99)}} & 82.84 {\color{Green} \footnotesize{(+0.18)}} & 88.70 {\color{Green} \footnotesize{(+0.94)}} & 79.57 {\color{Red} \footnotesize{(-2.03)}} & \textbf{90.83} {\color{Green} \footnotesize{(+3.64)}} & \textbf{83.29} {\color{Green} \footnotesize{(+0.64)}} \\ \midrule
\multirow{7}{*}{\textit{Two-stage}} 
& Gemma 3 4B & 88.29 {\color{Green} \footnotesize{(+1.90)}} & 82.43 {\color{Red} \footnotesize{(-0.23)}} & \textbf{91.08} {\color{Green} \footnotesize{(+3.32)}} & 79.82 {\color{Red} \footnotesize{(-1.78)}} & 89.36 {\color{Green} \footnotesize{(+2.17)}} & 82.90 {\color{Green} \footnotesize{(+0.25)}} \\
& Gemma 3 27B & 89.62 {\color{Green} \footnotesize{(+3.23)}} & 83.03 {\color{Green} \footnotesize{(+0.37)}} & 90.83 {\color{Green} \footnotesize{(+3.07)}} & 81.92 {\color{Green} \footnotesize{(+0.32)}} & 90.55 {\color{Green} \footnotesize{(+3.36)}} & 83.15 {\color{Green} \footnotesize{(+0.50)}} \\
& Claude 3.5 Haiku & 89.53 {\color{Green} \footnotesize{(+3.14)}} & 82.69 {\color{Green} \footnotesize{(+0.03)}} & 90.20 {\color{Green} \footnotesize{(+2.44)}} & 82.51 {\color{Green} \footnotesize{(+0.91)}} & 90.56 {\color{Green} \footnotesize{(+3.37)}} & 82.97 {\color{Green} \footnotesize{(+0.32)}} \\
& Claude Sonnet 4 & 89.85 {\color{Green} \footnotesize{(+3.46)}} & 83.45 {\color{Green} \footnotesize{(+0.79)}} & 90.56 {\color{Green} \footnotesize{(+2.80)}} & 82.94 {\color{Green} \footnotesize{(+1.34)}} & 90.76 {\color{Green} \footnotesize{(+3.57)}} & 83.22 {\color{Green} \footnotesize{(+0.57)}} \\
& Claude Sonnet 4.5 & \textbf{89.93} {\color{Green} \footnotesize{(+3.54)}} & \textbf{84.50} {\color{Green} \footnotesize{(+1.84)}} & 90.80 {\color{Green} \footnotesize{(+3.04)}} & \textbf{83.19} {\color{Green} \footnotesize{(+1.59)}} & 90.81 {\color{Green} \footnotesize{(+3.62)}} & 83.27 {\color{Green} \footnotesize{(+0.62)}} \\
& GPT OSS 20B & 89.10 {\color{Green} \footnotesize{(+2.71)}} & 82.67 {\color{Green} \footnotesize{(+0.01)}} & 87.98 {\color{Green} \footnotesize{(+0.22)}} & 79.06 {\color{Red} \footnotesize{(-2.54)}} & 89.56 {\color{Green} \footnotesize{(+2.37)}} & 82.10 {\color{Red} \footnotesize{(-0.55)}} \\
& GPT OSS 120B & 89.18 {\color{Green} \footnotesize{(+2.79)}} & 82.76 {\color{Green} \footnotesize{(+0.10)}} & 89.63 {\color{Green} \footnotesize{(+1.87)}} & 79.52 {\color{Red} \footnotesize{(-2.08)}} & 88.37 {\color{Green} \footnotesize{(+1.18)}} & 81.29 {\color{Red} \footnotesize{(-1.36)}} \\

\midrule[0.03em]
\multirow{2}{*}{\textbf{Method}} & \multirow{2}{*}{\textbf{Model}} & \multicolumn{2}{c}{\textsc{EN} $\to$ \textsc{ES}} & \multicolumn{2}{c}{\textsc{EN} $\to$ \textsc{IT}} & \multicolumn{2}{c}{\textsc{EN} $\to$ \textsc{JP}} \\ \cmidrule(lr){3-4} \cmidrule(lr){5-6} \cmidrule(lr){7-8}
& & \textbf{\textsc{Comet}} & \textbf{\textsc{CometKiwi}} & \textbf{\textsc{Comet}} & \textbf{\textsc{CometKiwi}} & \textbf{\textsc{Comet}} & \textbf{\textsc{CometKiwi}} \\ \midrule
\textit{No post-edit} & -- & 83.01 & 82.02 & 87.36 & \textbf{84.28} & 89.08 & \textbf{84.79} \\ \midrule
\multirow{1}{*}{\textit{One-stage}} & Claude Opus 4.5 & 89.59 {\color{Green} \footnotesize{(+6.58)}} & 81.73 {\color{Red} \footnotesize{(-0.29)}} & 91.25 {\color{Green} \footnotesize{(+3.89)}} & 82.77 {\color{Red} \footnotesize{(-1.51)}} & 93.48 {\color{Green} \footnotesize{(+4.40)}} & 83.96 {\color{Red} \footnotesize{(-0.83)}} \\ \midrule
\multirow{7}{*}{\textit{Two-stage}} 
& Gemma 3 4B & 83.73 {\color{Green} \footnotesize{(+0.72)}} & 80.16 {\color{Red} \footnotesize{(-1.86)}} & 89.36 {\color{Green} \footnotesize{(+2.00)}} & 82.11 {\color{Red} \footnotesize{(-2.17)}} & 89.73 {\color{Green} \footnotesize{(+0.65)}} & 82.52 {\color{Red} \footnotesize{(-2.27)}} \\
& Gemma 3 27B & 89.17 {\color{Green} \footnotesize{(+6.16)}} & 81.59 {\color{Red} \footnotesize{(-0.43)}} & 90.72 {\color{Green} \footnotesize{(+3.36)}} & 82.45 {\color{Red} \footnotesize{(-1.83)}} & 91.59 {\color{Green} \footnotesize{(+2.51)}} & 82.92 {\color{Red} \footnotesize{(-1.87)}} \\
& Claude 3.5 Haiku & 90.35 {\color{Green} \footnotesize{(+7.34)}} & 81.75 {\color{Red} \footnotesize{(-0.27)}} & 90.64 {\color{Green} \footnotesize{(+3.28)}} & 82.10 {\color{Red} \footnotesize{(-2.18)}} & 92.43 {\color{Green} \footnotesize{(+3.35)}} & 83.06 {\color{Red} \footnotesize{(-1.73)}} \\
& Claude Sonnet 4 & 90.91 {\color{Green} \footnotesize{(+7.90)}} & 82.05 {\color{Green} \footnotesize{(+0.03)}} & 91.02 {\color{Green} \footnotesize{(+3.66)}} & 82.57 {\color{Red} \footnotesize{(-1.71)}} & 92.82 {\color{Green} \footnotesize{(+3.74)}} & 83.79 {\color{Red} \footnotesize{(-1.00)}} \\
& Claude Sonnet 4.5 & \textbf{90.93} {\color{Green} \footnotesize{(+7.92)}} & \textbf{82.12} {\color{Green} \footnotesize{(+0.10)}} & \textbf{91.05} {\color{Green} \footnotesize{(+3.69)}} & 82.63 {\color{Red} \footnotesize{(-1.65)}} & \textbf{92.87} {\color{Green} \footnotesize{(+3.79)}} & 83.80 {\color{Red} \footnotesize{(-0.99)}} \\
& GPT OSS 20B & 87.30 {\color{Green} \footnotesize{(+4.29)}} & 79.00 {\color{Red} \footnotesize{(-3.02)}} & 89.79 {\color{Green} \footnotesize{(+2.43)}} & 81.29 {\color{Red} \footnotesize{(-2.99)}} & 91.71 {\color{Green} \footnotesize{(+2.63)}} & 82.71 {\color{Red} \footnotesize{(-2.08)}} \\
& GPT OSS 120B & 87.44 {\color{Green} \footnotesize{(+4.43)}} & 78.99 {\color{Red} \footnotesize{(-3.03)}} & 88.82 {\color{Green} \footnotesize{(+1.46)}} & 80.72 {\color{Red} \footnotesize{(-3.56)}} & 90.83 {\color{Green} \footnotesize{(+1.75)}} & 82.12 {\color{Red} \footnotesize{(-2.67)}} \\
\bottomrule
\end{tabular}
\end{adjustbox}
\caption{\textsc{Comet}-22 and \textsc{CometKiwi} scores of $y^{(0)}$ and $y^{(1)}$ across all considered LLMs, using \textbf{P3} as the evaluator prompting strategy}
\label{tab:full}
\end{table*}

The complete results across all six language directions, reported in Table~\ref{tab:full}, confirm that our findings from \textsc{EN} $\to$ \textsc{DE} and \textsc{EN} $\to$ \textsc{CN} generalize broadly. Evaluator-guided post-editing consistently improves \textsc{Comet}-22 scores over the no-post-edit baseline regardless of target language, and the two-stage design outperforms the one-stage baseline in the majority of cases across all directions. \textsc{EN} $\to$ \textsc{FR} is a notable anomaly where one-stage post-edit yields the best result over all two-stage results, although the gap between the one-stage performance and the best two-stage performance (Claude Sonnet 4.5) is minimal.

Claude Sonnet 4.5 remains the strongest post-editor overall, achieving the best or near-best \textsc{Comet}-22 scores in most language directions, while the Claude family models in general demonstrate strong and consistent gains across European languages. The GPT OSS models yield competitive \textsc{Comet}-22 improvements for European target languages but show more variable \textsc{CometKiwi} behavior, echoing the pattern observed in the main results. These results further support the recommendation that practitioners should experiment with a wide range of models for their specific languages of interest and domains prior to selecting a post-editor for deployment.

\newpage
\section{Ablation Studies and Further Investigation of the System}

\subsection{Varying Evaluator LLMs}

In the main experiments, we reserved Claude Opus 4.5 as our evaluator LLM. A reasonable question that arises concerns how the system's performance shifts when we utilize different evaluators. To investigate this effect, we conducted an ablation study on \textsc{EN} $\to$ \textsc{FR}, varying the evaluator model while keeping the post-editor (Claude Sonnet 4.5) and the prompt (\textbf{P3}) fixed. The results are as follows:

\begin{table}[H]
\centering
\begin{tabular}{ccc}
\toprule
\textbf{Evaluator LLM} & \textbf{\textsc{Comet}} & \textbf{\textsc{CometKiwi}} \\ \midrule
No post-edit & 87.19 & 82.65 \\ \midrule
Claude 3.5 Haiku & 88.28 {\color{Green} \footnotesize{(+1.09)}} & \textbf{83.97} {\color{Green} \footnotesize{(+1.32)}} \\
Claude Sonnet 4 & 89.80 {\color{Green} \footnotesize{(+2.61)}} & 83.59 {\color{Green} \footnotesize{(+0.94)}} \\
Claude Sonnet 4.5 & 90.37 {\color{Green} \footnotesize{(+3.18)}} & 83.10 {\color{Green} \footnotesize{(+0.45)}} \\
Claude Opus 4.5 & \textbf{90.81} {\color{Green} \footnotesize{(+3.62)}} & 83.27 {\color{Green} \footnotesize{(+0.62)}} \\
\bottomrule
\end{tabular}
\caption{\textsc{Comet}-22 and \textsc{CometKiwi} scores of $y^{(1)}$ upon varying the evaluator LLM}
\label{tab:varying-evaluator}
\end{table}

We observe that even the smallest and most cost-efficient evaluator (Claude 3.5 Haiku) achieves a \textsc{Comet}-22 score of 88.28, a $+$1.09 improvement over the no post-edit baseline. This gain is obtained using an evaluator that is substantially cheaper than Claude Opus 4.5, demonstrating that the framework does not critically depend on the most powerful evaluator to deliver improvements. As the evaluator model scales up, \textsc{Comet}-22 increases monotonically, indicating that stronger evaluators do contribute additional gains, but the framework remains effective across the full range of evaluator capabilities.

\subsection{Comparison Against GEMBA-MQM}

For our main set of experiments as displayed in Table~\ref{tab:main-de-cn}, we tested one-stage framework paired with \textbf{P3} as the evaluator prompt. For a fair comparison, we conducted additional experiments to validate our approach against GEMBA-MQM as one-stage post-edit process, where the model is asked to both diagnose and correct errors in a single inference call. The results on \textsc{EN} $\to$ \textsc{DE} are as follows:

\begin{table}[H]
\centering
\begin{tabular}{cccc}
\toprule
\textbf{Method} & \textbf{Model} & \textbf{\textsc{Comet}} & \textbf{\textsc{CometKiwi}} \\ \midrule
\textit{No post-edit} & Claude 3.5 Sonnet & 86.39 & 82.66 \\ \midrule
\textit{One-stage} & \multirow{2}{*}{Claude Opus 4.5} & \multirow{2}{*}{86.51 {\color{Green} \footnotesize{(+0.12)}}} & \multirow{2}{*}{82.88 {\color{Green} \footnotesize{(+0.22)}}} \\ 
{\small (GEMBA-MQM)} & & & \\ \midrule
\multirow{2}{*}{\textit{Two-stage}} & Claude Opus 4.5 $+$ & \multirow{2}{*}{\textbf{89.93} {\color{Green} \footnotesize{(+3.54)}}} & \multirow{2}{*}{\textbf{84.50} {\color{Green} \footnotesize{(+1.84)}}} \\
& Claude Sonnet 4.5 & & \\
\bottomrule
\end{tabular}
\caption{\textsc{Comet}-22 and \textsc{CometKiwi} scores for $y^{(0)}$ and $y^{(1)}$ for one-stage GEMBA-MQM (\textit{i.e.} with prompt \textbf{P1}) and our two-stage framework}
\label{tab:gemba-mqm-comparison}
\end{table}

The one-stage GEMBA-MQM baseline yields only marginal improvements over the original translation ($+$0.12 \textsc{Comet}-22 and $+$0.22 \textsc{CometKiwi}), despite identifying and attempting to correct 44.8\% of segments. This is consistent with prior findings that LLMs struggle to simultaneously diagnose and repair their own outputs in a single pass. In contrast, our two-stage framework achieves substantially larger gains ($+$3.54 \textsc{Comet}-22, $+$1.84 \textsc{CometKiwi}), validating the core thesis of our proposed framework that decoupling evaluation from post-editing enables more effective translation refinement.

\newpage
\section{Latency Analysis}

The following table provides the latency statistic involved in the post-edit procedure, computed over 50 segments:

\begin{table}[H]
\centering
\begin{tabular}{lccccc}
\toprule
\textbf{Stage} & \textbf{Mean} & \textbf{Median} & \textbf{P75} & \textbf{P95} & \textbf{Stdev} \\ \midrule
Evaluator (Claude Opus 4.5) & 14.34 & 7.80 & 12.54 & 54.38 & 17.38 \\
Post-editor (Claude Sonnet 4.5) & 3.30 & 3.10 & 4.15 & 5.04 & 1.08 \\
End-to-end (single round) & 17.64 & 10.90 & 16.69 & 59.41 & 17.41 \\
\bottomrule
\end{tabular}
\caption{Latency statistics (in seconds) for each stage of the two-stage pipeline, measured over 50 segments}
\label{tab:latency}
\end{table}

Based on these numbers, we contend that, while the approach may be limited in situations that require instantaneous edits (such as conversational settings), it would still be suitable for situations that can tolerate slight latency for offline evaluation.

In domain-specific machine translation, the content is often highly repetitive. For e-commerce, for example, segments contain a lot of product descriptions, policy text, and shipping notices. As such, a cache keyed on source segment embeddings could bypass the evaluator entirely for segments similar to previously diagnosed ones, delivering an additional advantage of reduced costs. In our dataset, we also observe significant lexical overlap across segments, suggesting that the cache hit rates would be very high in practice.

\newpage
\section{List of LLMs and Dataset Details} \label{sec:model-dataset-details}

We compose our universe of post-editor LLMs with models of varying providers and sizes, to have a glimpse of the effect of model diversity and size on the quality of post-edits. We focus on models that support batch inference on Amazon Bedrock as of February 2026. We further note that, for certain models, the information regarding the number of active parameters is not publicly released, in which case we mark the relevant cell with $(*)$. The unit of context window and max output is tokens; the ``Pricing'' column lists the input and output prices per million tokens with real-time inference on Amazon Bedrock.

Tables~\ref{tab:model-details} and \ref{tab:dataset-details} summarize the LLMs and data splits for each language direction we used for the experiments, respectively:

\begin{table}[h]
\centering
\begin{adjustbox}{width=\textwidth}
\begin{tabular}{l c c c c c l}
\toprule
& \textbf{Model} & \textbf{\# Params} & \textbf{Bedrock ID} & \textbf{Context / Max Output} & \textbf{Pricing} \\ \midrule
\multicolumn{2}{l}{\textbf{Google}\,\,\googlelogo} \\[2pt]
1 & \multicolumn{1}{l}{Gemma 3 4B IT} & 4 billion & \texttt{google.gemma-3-4b-it} & 128,000 / 8,192 & \$0.04 / \$0.08 \\
2 & \multicolumn{1}{l}{Gemma 3 27B PT} & 27 billion & \texttt{google.gemma-3-27b-it} & 128,000 / 8,192 & \$0.23 / \$0.38 \\ \cmidrule(lr){1-6}
\multicolumn{2}{l}{\textbf{Anthropic}\,\,\anthropiclogo} \\[2pt]
1 & \multicolumn{1}{l}{Claude 3.5 Haiku} & $(*)$ & \texttt{anthropic.claude3-5-haiku-20241022-v1:0} & 200,000 / 8,192 & \$0.8 / \$4 \\
2 & \multicolumn{1}{l}{Claude Sonnet 4} & $(*)$ & \texttt{anthropic.claude-sonnet-4-20250514-v1:0} & 200,000 / 8,192 & \$3 / \$15 \\
3 & \multicolumn{1}{l}{Claude Sonnet 4.5} & $(*)$ & \texttt{anthropic.claude-sonnet-4-5-20250929-v1:0} & 200,000 / 64,000 & \$3 / \$15 \\
4 & \multicolumn{1}{l}{Claude Opus 4.5} & $(*)$ & \texttt{anthropic.claude-opus-4-5-20251101-v1:0} & 200,000 / 64,000 & \$5 / \$25 \\ \cmidrule(lr){1-6}
\multicolumn{2}{l}{\textbf{OpenAI}\,\,\openailogo} \\[2pt]
1 & \multicolumn{1}{l}{GPT OSS 20B} & 20 billion & \texttt{openai.gpt-oss-20b-1:0} & 128,000 / 16,000 & \$0.07 / \$0.3 \\
2 & \multicolumn{1}{l}{GPT OSS 120B} & 120 billion & \texttt{openai.gpt-oss-120b-1:0} & 128,000 / 16,000 & \$0.15 / \$0.6 \\
\bottomrule
\end{tabular}
\end{adjustbox}
\caption{Model details}
\label{tab:model-details}
\end{table}

\begin{table}[h]
\centering
\begin{adjustbox}{width=0.7\textwidth}
\begin{tabular}{l c c c cc}
\toprule
& \multirow{2}{*}{\textbf{Name}} & \multirow{2}{*}{\textbf{Domain}} & \multirow{2}{*}{\textbf{Language Direction}} & \multicolumn{2}{c}{\textbf{Split Size}} \\
\cmidrule(lr){5-6}
& & & & \textbf{KB} & \textbf{Test} \\ \midrule
1 & \multirow{6}{*}{\makecell{Translation Memory \\ (TM)}} & \multirow{6}{*}{e-commerce} & \textsc{EN} $\to$ \textsc{DE} & 135,000 & 5,000 \\ 
2 & &  & \textsc{EN} $\to$ \textsc{IT} & 140,000 & 5,000 \\
3 & &  & \textsc{EN} $\to$ \textsc{ES} & 184,000 & 5,000 \\
4 & &  & \textsc{EN} $\to$ \textsc{FR} & 141,000 & 5,000 \\
5 & &  & \textsc{EN} $\to$ \textsc{CN} & 170,000 & 5,000 \\
6 & &  & \textsc{EN} $\to$ \textsc{JP} & 152,000 & 5,000 \\
\bottomrule
\end{tabular}
\end{adjustbox}
\caption{Dataset details}
\label{tab:dataset-details}
\end{table}




\newpage
\section{Prompts} \label{sec:evaluator-and-post-editor-prompt}

\subsection{Evaluator Prompts}

\begin{tcolorbox}[width=1\textwidth,colback={white},title={\textbf{Evaluator Prompt} (GEMBA-MQM + Rubric + Edit Contract + RAG)},colbacktitle={black},coltitle=white,colframe={black},breakable]
\parskip=1pt

{\footnotesize

You are an annotator for the quality of machine translation. Your task is to identify errors and assess the quality of the translation, grounded in domain-specific knowledge provided via retrieved exemplars.

\vspace{5pt}
Domain exemplars (high-quality reference translations from the domain):
\begin{verbatim}
{domain_exemplars}
\end{verbatim}

\vspace{5pt}
\texttt{\{source\_lang\}} source:
\begin{verbatim}
{source_segment}
\end{verbatim}

\vspace{5pt}
\texttt{\{target\_lang\}} translation:
\begin{verbatim}
{target_segment}
\end{verbatim}

\vspace{5pt}
Based on the source segment, machine translation, and domain exemplars surrounded with triple backticks, identify error types in the translation and classify them.

\vspace{5pt}
Grounding Rules:
\begin{itemize}[label=--]
    \item Terminology Consistency: If the translation uses a term that contradicts a term used in the domain exemplars, flag it as a Terminology error.
    \item Style Alignment: Use the exemplars to determine the appropriate register (e.g., formal vs. informal) for this domain.
    \item Evidence: For every identified error, if applicable, reference the specific ``\texttt{evidence\_id}'' from the domain exemplars that justifies your diagnosis.
\end{itemize}

\vspace{5pt}
ERROR RUBRIC:
\vspace{5pt}
Accuracy errors (meaning transfer issues):
\begin{itemize}[label=--]
    \item \texttt{accuracy/addition}: Translation includes content not present in source (e.g., adding explanatory text)
    \item \texttt{accuracy/mistranslation}: Source meaning incorrectly rendered (e.g., ``hot'' $\rightarrow$ ``cold'', negation flipped)
    \item accuracy/omission: Source content missing from translation (e.g., clause or word dropped)
    \item \texttt{accuracy/untranslated\_text}: Source text left untranslated when translation expected
\end{itemize}

\vspace{5pt}
Fluency errors (target language quality):
\begin{itemize}[label=--]
    \item \texttt{fluency/grammar}: Grammatical errors (e.g., subject-verb disagreement, wrong tense)
    \item \texttt{fluency/spelling}: Spelling errors or typos
    \item \texttt{fluency/punctuation}: Incorrect punctuation or typography for target locale
    \item \texttt{fluency/character\_encoding}: Display issues (e.g., mojibake, garbled characters)
    \item \texttt{fluency/inconsistency}: Same term translated differently without justification
    \item \texttt{fluency/register}: Wrong formality level (e.g., informal ``du'' in formal document)
\end{itemize}

\vspace{5pt}
Locale convention errors (formatting for target region):
\begin{itemize}[label=--]
    \item \texttt{locale/currency\_format}: Currency not formatted per target locale (e.g., ``\$100'' vs ``100 €'')
    \item \texttt{locale/date\_format}: Date not formatted per target locale (e.g., ``12/31/2024'' vs ``31.12.2024'')
    \item \texttt{locale/number\_format}: Numbers not formatted per target locale (e.g., ``1,000.50'' vs ``1.000,50'')
    \item \texttt{locale/time\_format}: Time not formatted per target locale
    \item \texttt{locale/name\_format}: Names not ordered per target conventions
    \item \texttt{locale/telephone\_format}: Phone numbers not formatted correctly
\end{itemize}

\vspace{5pt}
Terminology errors (domain-specific term issues):
\begin{itemize}[label=--]
    \item \texttt{terminology/inappropriate\_for\_context}: Term technically correct but wrong for domain
    \item \texttt{terminology/inconsistent\_use}: Domain term differs from exemplars (cite \texttt{evidence\_id})
\end{itemize}

\vspace{5pt}
Style errors:
\begin{itemize}[label=--]
    \item \texttt{style/awkward}: Grammatically correct but unnatural phrasing
\end{itemize}

\vspace{5pt}
Other:
\begin{itemize}[label=--]
    \item \texttt{other/non\_translation}: Output is not a translation (e.g., refusal, error message)
    \item \texttt{other/no\_error}: No errors found
\end{itemize}

\vspace{5pt}
SEVERITY LEVELS:
\begin{itemize}[label=--]
    \item critical: Errors that inhibit comprehension or could cause harm
    \item major: Errors that disrupt flow but text is still understandable
    \item minor: Technical errors that do not disrupt flow or hinder comprehension
\end{itemize}

\vspace{5pt}
CHAIN OF THOUGHT:
\vspace{5pt}
Before providing the final XML output, think through your evaluation step by step:
\begin{enumerate}
    \item First, compare source and translation for accuracy (additions, omissions, mistranslations)
    \item Then check if exemplars establish terminology patterns relevant to this segment
    \item Check fluency issues (grammar, spelling, punctuation)
    \item Check locale formatting if applicable
    \item Assign severity based on user impact
\end{enumerate}

\vspace{5pt}
Requirements:
\begin{enumerate}
    \item Identify each error with its specific category and subcategory
    \item Assign appropriate severity level to each error
    \item Provide brief explanation for each identified error
    \item Reference \texttt{evidence\_id} from domain exemplars when applicable
    \item If no errors found, respond with ``\texttt{no-error}''
\end{enumerate}

\vspace{5pt}
OUTPUT FORMAT:

\vspace{5pt}
Return ONLY the XML structure below. Do NOT include explanations, notes, preambles, or markdown code blocks.


\vspace{5pt}
For translations WITH errors (include one \texttt{<error>} block per error found):
\begin{verbatim}
<evaluation>
  <has_errors>true</has_errors>
  <errors>
\end{verbatim}
\vspace{2pt}
\begin{minipage}[c]{0.63\textwidth} 
\begin{verbatim}
    <error>
      <category>CATEGORY/SUBCATEGORY</category>
      <severity>critical|major|minor</severity>
      <span>ERRONEOUS_TEXT</span>
      <explanation>BRIEF_EXPLANATION</explanation>
      <evidence_id>EXEMPLAR_ID_IF_APPLICABLE</evidence_id>
      <suggested_edit>CORRECTED_TEXT_FOR_SPAN</suggested_edit>
    </error>
\end{verbatim}
\end{minipage}%
\hfill
\begin{minipage}[c]{0.37\textwidth} 
    \centering
    \begin{tabular}{@{}c@{}}
        \hspace{20pt} {\footnotesize \color{gray} $\rceil$} \\[-2pt] 
        \begin{minipage}{2.8cm} 
        \vspace{0.5cm}
            \raggedright \scriptsize \itshape \color{gray}
            Repeat \texttt{<error>} block \\ for each additional error
            \vspace{0.5cm}
        \end{minipage} \\[-2pt] 
        \hspace{20pt} {\footnotesize \color{gray} $\rfloor$}
    \end{tabular}
\end{minipage}
\begin{verbatim}
  </errors>
</evaluation>
\end{verbatim}

\vspace{5pt}
EDIT CONTRACT:
\begin{itemize}[label=--]
    \item \texttt{suggested\_edit} must be a direct replacement for the text in \texttt{<span>}
    \item For omissions, provide the missing text that should be inserted
    \item For additions, use empty string or ``\texttt{[DELETE]}'' to indicate removal
    \item Keep edits minimal - only fix the identified error, preserve surrounding context
    \item If no clear fix exists (e.g., \texttt{style/awkward}), provide the most natural alternative
\end{itemize}

\vspace{5pt}
For translations with NO errors:
\begin{verbatim}
<evaluation>
  <has_errors>false</has_errors>
  <errors></errors>
</evaluation>
\end{verbatim}

}

\end{tcolorbox}

\subsection{Post-Editor Prompts}

\begin{tcolorbox}[width=1\textwidth,colback={white},title={\textbf{Post-Editor Prompt for $C_1$}},colbacktitle={black},coltitle=white,colframe={black},breakable]
\parskip=1pt

{\footnotesize

You are an expert translation post-editor. Your task is to perform EVIDENCE-TARGETED post-editing, correcting ONLY the specific errors identified by the quality evaluator. Do not make unnecessary changes to parts of the translation that were not flagged.

\vspace{5pt}
\texttt{\{source\_lang\}} source:
\begin{verbatim}
{source_text}
\end{verbatim}

\vspace{5pt}
\texttt{\{target\_lang\}} translation:
\begin{verbatim}
{translated_text}
\end{verbatim}

\vspace{5pt}
EVALUATION RESULTS:

\vspace{5pt}
Has Errors: \texttt{\{has\_errors\}}

\vspace{5pt}
\texttt{\{errors\_section\}}

\vspace{5pt}
INSTRUCTIONS:

\vspace{5pt}
\begin{enumerate}
\item \textbf{If no errors were identified} (\texttt{has\_errors: false}):
\begin{enumerate}[label=--]
    \item Return the original translation UNCHANGED
    \item Do not attempt to "improve" correct translations
\end{enumerate}
\item \textbf{If errors were identified} (`\texttt{has\_errors: true}'):
\begin{enumerate}[label=--]
    \item Review each error's category, severity, span, and explanation
    \item Make MINIMAL, TARGETED corrections only for the identified errors
    \item Preserve ALL non-erroneous parts of the translation exactly as they are
    \item Prioritize critical errors, then major, then minor
    \item Ensure corrections maintain consistency with the rest of the translation
\end{enumerate}
\item \textbf{Correction Guidelines by Error Category}:
\begin{enumerate}[label=--]
    \item \textbf{Accuracy errors}: Fix mistranslations, add omitted content, remove additions, translate untranslated text
    \item \textbf{Fluency errors}: Fix grammar, spelling, punctuation while preserving meaning
    \item \textbf{Terminology errors}: Use suggested, appropriate domain-specific terms based on error detail
    \item \textbf{Style errors}: Improve awkward phrasing minimally
    \item \textbf{Locale convention errors}: Adjust formats to target locale standards
\end{enumerate}
\end{enumerate}

\vspace{5pt}
IMPORTANT OUTPUT FORMAT:

\vspace{5pt}
Return ONLY the XML structure below. Do NOT include explanations, notes, preambles, or markdown code blocks outside the XML.

\vspace{5pt}
For translations that NEEDED corrections:
\begin{verbatim}
<correction>
  <corrected>true</corrected>
  <corrected_translation>FULL_CORRECTED_TRANSLATION_HERE</corrected_translation>
  <changes>
\end{verbatim}

\noindent
\begin{minipage}[c]{0.63\textwidth} %
\begin{verbatim}
    <change>
      <original_span>ORIGINAL_ERRONEOUS_TEXT</original_span>
      <corrected_span>CORRECTED_TEXT</corrected_span>
      <error_addressed>CATEGORY/SUBCATEGORY</error_addressed>
      <rationale>BRIEF_EXPLANATION_OF_FIX</rationale>
    </change>
\end{verbatim}
\end{minipage}%
\hfill
\begin{minipage}[c]{0.37\textwidth} 
    \centering
    \begin{tabular}{@{}c@{}}
        \hspace{20pt} {\footnotesize \color{gray} $\rceil$} \\[-2pt] 
        \begin{minipage}{2.8cm} 
        \vspace{0.2cm}
            \raggedright \scriptsize \itshape \color{gray}
            Repeat \texttt{<change>} block for each correction made
            \vspace{0.2cm}
        \end{minipage} \\[-2pt] 
        \hspace{20pt} {\footnotesize \color{gray} $\rfloor$}
    \end{tabular}
\end{minipage}
\begin{verbatim}
  </changes>
</correction>
\end{verbatim}

\vspace{5pt}
For translations with NO errors (unchanged):
\begin{verbatim}
<correction>
  <corrected>false</corrected>
  <corrected_translation>ORIGINAL_TRANSLATION_UNCHANGED</corrected_translation>
  <changes></changes>
</correction>
\end{verbatim}

}
\end{tcolorbox}

\begin{tcolorbox}[width=1\textwidth,colback={white},title={\textbf{Post-Editor Prompt for $C_2$}},colbacktitle={black},coltitle=white,colframe={black},breakable]
\parskip=1pt

{\footnotesize

You are an expert translation verifier and post-editor for \texttt{\{source\_lang\}} to \texttt{\{target\_lang\}} translation.

\vspace{5pt}
Your task is to verify whether a previous correction attempt (Post-editor 1) has resolved all identified translation errors, and if not, fix the remaining issues.

\vspace{5pt}
\#\# INPUT

\vspace{5pt}
Source text (\texttt{\{source\_lang\}}):

\vspace{5pt}
\texttt{\{source\_text\}}

\vspace{5pt}
Original translation (before correction):

\vspace{5pt}
\texttt{\{original\_translation\}}

\vspace{5pt}
Corrected translation (after Post-editor 1):

\vspace{5pt}
\texttt{\{corrected\_translation\}}

\vspace{5pt}
Original errors identified by evaluator:

\vspace{5pt}
\texttt{\{errors\_section\}}

\vspace{5pt}
Changes made by Post-editor 1:

\vspace{5pt}
\texttt{\{changes\_section\}}

\vspace{5pt}
\#\# YOUR TASK

\vspace{5pt}
\begin{enumerate}
    \item \textbf{Verify each error}: For each error identified by the evaluator, determine if Post-editor 1's changes have adequately addressed it.
    \item \textbf{Identify unresolved errors}: List any errors that remain unresolved or were only partially fixed.
    \item \textbf{Apply additional corrections}: If there are unresolved errors, apply the necessary corrections to fully resolve them.
\end{enumerate}

\vspace{5pt}
\#\# OUTPUT FORMAT

\vspace{5pt}
Return ONLY the XML structure below. Do NOT include explanations, notes, preambles, or markdown code blocks outside the XML.

  


\begin{verbatim}
<verification>
  <all_errors_resolved>true/false</all_errors_resolved>
  <error_status>
\end{verbatim}

\noindent
\begin{minipage}[c]{0.83\textwidth} 
\begin{verbatim}
    <error>
      <original_error>Description of the original error</original_error>
      <status>resolved/partially_resolved/unresolved</status>
      <assessment>Explanation of whether and how it was fixed</assessment>
    </error>
\end{verbatim}
\end{minipage}%
\hfill
\begin{minipage}[c]{0.17\textwidth} 
    \begin{tabular}{@{}c@{}}
        \hspace{-40pt} {\footnotesize \color{gray} $\rceil$} \\[-2pt] 
        \begin{minipage}{2.8cm}
            \vspace{0.2cm}
            \raggedright \scriptsize \itshape \color{gray}
            Repeat \texttt{<error>} block for each original error
            \vspace{0.2cm}
        \end{minipage} \\[-2pt] 
        \hspace{-40pt} {\footnotesize \color{gray} $\rfloor$}
    \end{tabular}
\end{minipage}

\begin{verbatim}
  </error_status>
  <additional_correction_needed>true/false</additional_correction_needed>
  <final_translation>
    The final corrected translation (either the Post-editor 1 output if all resolved, 
    or your further corrected version)
  </final_translation>
\end{verbatim}

\noindent
\begin{minipage}[c]{0.85\textwidth} 
\vspace{2pt}
\begin{verbatim}
  <additional_changes>
    <change>
      <original_span>Text that was changed</original_span>
      <corrected_span>New corrected text</corrected_span>
      <error_addressed>Which unresolved error this fixes</error_addressed>
      <rationale>Why this correction is needed</rationale>
    </change>
  </additional_changes>
\end{verbatim}
\end{minipage}%
\hfill
\begin{minipage}[c]{0.15\textwidth} 
    \begin{tabular}{@{}c@{}}
        \hspace{-40pt} {\footnotesize \color{gray} $\rceil$} \\[-2pt] 
        \begin{minipage}{2.8cm}
            \vspace{0.4cm}
            \raggedright \scriptsize \itshape \color{gray}
            Include only if \\ additional corrections \\ are needed
            \vspace{0.4cm}
        \end{minipage} \\[-2pt] 
        \hspace{-40pt} {\footnotesize \color{gray} $\rfloor$}
    \end{tabular}
\end{minipage}
\begin{verbatim}
</verification>
\end{verbatim}

\vspace{5pt}
\#\# GUIDELINES

\vspace{5pt}
\begin{itemize}[label=--]
    \item Be thorough: Check each original error against the corrected translation
    \item Be conservative: Only make additional changes if truly necessary
    \item Preserve good corrections: Don't undo successful corrections from Post-editor 1
    \item Focus on unresolved issues: Don't introduce new changes for issues not in the original error list
    \item If all errors are resolved, set \texttt{all\_errors\_resolved} to true and return the Post-editor 1 output unchanged
\end{itemize}

}
\end{tcolorbox}

\subsection{Prompt for Translation}

\begin{tcolorbox}[width=1\textwidth,colback={white},title={\textbf{Prompt for Translation (\textsc{EN} $\to$ \textsc{CN} as an example)}},colbacktitle={black},coltitle=white,colframe={black},breakable]
\parskip=1pt

{\footnotesize
You are a professional translator. Your task is to translate the following English text into Chinese. Do not output anything other than a single Chinese translation for the text within \texttt{<translation>} and \texttt{</translation>} XML tag. Return exactly one XML element in the form: \texttt{<translation> ... </translation>}.

\vspace{5pt}
\noindent

\begin{minipage}[c]{0.77\textwidth}
The following are \texttt{\{num\_exemplars\}} exemplar translations of similar English texts:

\begin{verbatim}
<exemplar_1>
\end{verbatim}
$\quad$English: \texttt{\{english\_1\}}

$\quad$Chinese: \texttt{\{chinese\_1\}}
\begin{verbatim}
</exemplar_1>
...
\end{verbatim}
\end{minipage}%
\hfill
\begin{minipage}[c]{0.23\textwidth} 
    \begin{tabular}{@{}c@{}}
        \hspace{-30pt} {\footnotesize \color{gray} $\rceil$} \\[-2pt] 
        \begin{minipage}{2.8cm}
            \vspace{0.3cm}
            \raggedright \scriptsize \itshape \color{gray}
            This section is only included in the presence of exemplars
            \vspace{0.3cm}
        \end{minipage} \\[-2pt] 
        \hspace{-30pt} {\footnotesize \color{gray} $\rfloor$}
    \end{tabular}
\end{minipage}

\vspace{10pt}
Translate the following text:

\begin{verbatim}
<text>
  {source_text}
</text>
\end{verbatim}

}
\end{tcolorbox}

\end{document}